\documentclass[sip,biber]{now-journal} 
\usepackage[utf8]{inputenc}      
\usepackage{csquotes}
\usepackage{amssymb}
\usepackage{float}
\usepackage{multirow}
\usepackage{graphicx}

\usepackage{colortbl}
\usepackage[table]{xcolor}

\usepackage{microtype}
\usepackage{ragged2e}
\AtBeginBibliography{\justifying}

\newcommand{\rmnum}[1]{\romannumeral #1}

\title{Automatic Medical Report Generation: Methods and Applications}

\author[1*]{Li Guo}
\author[1]{Anas M. Tahir}
\author[1]{Dong Zhang}
\author[1]{Z. Jane Wang}
\author[1]{Rabab K. Ward}
\affil{Electrical and Computer Engineering Department, University of British Columbia, Vancouver, Canada}

\creditline*{Corresponding author: Li Guo, lguo@ece.ubc.ca.}

\begin{document}

\begin{abstract}
The increasing demand for medical imaging has surpassed the capacity of available radiologists, leading to diagnostic delays and potential misdiagnoses. Artificial intelligence (AI) techniques, particularly in automatic medical report generation (AMRG), offer a promising solution to this dilemma. This review comprehensively examines AMRG methods from 2021 to 2024. It (\rmnum{1}) presents solutions to primary challenges in this field, (\rmnum{2}) explores AMRG applications across various imaging modalities, (\rmnum{3}) introduces publicly available datasets, (\rmnum{4}) outlines evaluation metrics, (\rmnum{5}) identifies techniques that significantly enhance model performance, and (\rmnum{6}) discusses unresolved issues and potential future research directions. This paper aims to provide a comprehensive understanding of the existing literature and inspire valuable future research.
\end{abstract}

\section{Introduction}
\label{Introduction}
Automatic medical report generation (AMRG) is an emerging research area in artificial intelligence (AI) within the medical field \cite{pang2023survey, jing2017automatic}. It utilizes computer vision (CV) and natural language processing (NLP) to interpret medical images and generate descriptive, human-like reports. AMRG has been applied to various imaging modalities, including X-rays, computed tomography (CT) scans, magnetic resonance imaging (MRI), and ultrasound \cite{hussain2022modern, allaouzi2018automatic, abhisheka2023recent}. This technology has the potential to streamline the diagnostic process, alleviate the workload on radiologists, and enhance diagnostic accuracy.

Traditionally, the interpretation of medical images relies on trained radiologists, a labor-intensive and error-prone process \cite{bruno2015understanding, aljuaid2022survey, xue2018multimodal, alfarghaly2021automated}. In the US and UK, the number of radiologists is insufficient to meet the growing demand for imaging and diagnostics \cite{rosenkrantz2016us, rimmer2017radiologist}. In resource-poor regions, the scarcity of radiology services is even more severe \cite{idowu2020diagnostic, rosman2015imaging}. This shortage of radiologists leads to delays and backlogs in interpreting medical images. In 2015, approximately 330,000 patients in the UK waited more than 30 days for radiology reports \cite{mayor2015waiting}. Due to delayed reports, some urgent images have to be reviewed by emergency physicians. However, the discernible interpretation differences between emergency physicians and trained radiologists can lead to missed diagnoses and misdiagnoses \cite{gatt2003chest}. Additionally, reports written by professional radiologists exhibit a 3-5\% error rate and approximately 35\% uncertainty rate \cite{brady2017error, srinivasa2015malpractice, lindley2014communicating}. As workloads increase, the probability of errors by radiologists also rises \cite{fitzgerald2001error, krupinski2010long}. For instance, an American doctor was sued after failing to detect a case of breast cancer due to reading too many X-rays in one day \cite{berlin2000liability}. AMRG addresses these issues by providing a systematic approach to image interpretation, potentially improving diagnostic efficiency and accuracy.

In recent years, deep learning has made significant progress in image analysis, with convolutional neural networks (CNNs) and Transformers excelling in high-precision lesion detection and classification of medical conditions \cite{srinivasan2024hybrid, he2023transformers, leung2022deep}. NLP techniques translate visual information from medical images into natural language reports, covering imaging findings, diagnostic conclusions, and recommendations, thereby achieving seamless image-to-text conversion \cite{casey2021systematic, shin2017natural, garcia2023integrating}. Researchers have developed various AMRG methods by combining CNNs, Transformers, and NLP in an encoder-decoder architecture \cite{pang2023survey, aksoy2023radiology, beddiar2023automatic}.

Despite these advancements, this field still faces numerous challenges. Firstly, bridging the modal gap between image input and text output is a fundamental challenge for AMRG. Medical images contain complex information that must be accurately interpreted and translated into coherent text, requiring sophisticated algorithms to map visual patterns to medical terminology. Secondly, medical images exhibit unique visual deviations: lesion areas usually occupy a small portion of the image, leading to highly similar normal and abnormal images, necessitating AMRG systems to be more sensitive to fine-grained differences than general image captioning models \cite{zhao2023normal, bria2020addressing, han2021madgan}. Thirdly, medical reports are long texts with high clinical professionalism and accuracy, placing higher standards on the quality of the generated texts, which demands advanced NLP techniques to handle detailed and precise medical documentation \cite{kaur2022methods, abdelrahman2014medical, estopa2020terminology}. Finally, medical datasets are limited and noisy; datasets like MIMIC-CXR (0.22M) \cite{johnson2019mimic} and IU-Xray (4k) \cite{demner2016preparing} are smaller than image recognition datasets like ImageNet (14M) \cite{deng2009imagenet} and image captioning datasets like Conceptual Captions (3.3M) \cite{sharma2018conceptual}, limiting model training effectiveness. Furthermore, noise in medical reports, such as temporal information, can confuse models and lead to inaccuracies or hallucinations \cite{ramesh2022improving, bannur2023learning}.

\begin{figure}[htbp]
    \centering
    \includegraphics[width=1.0\textwidth]{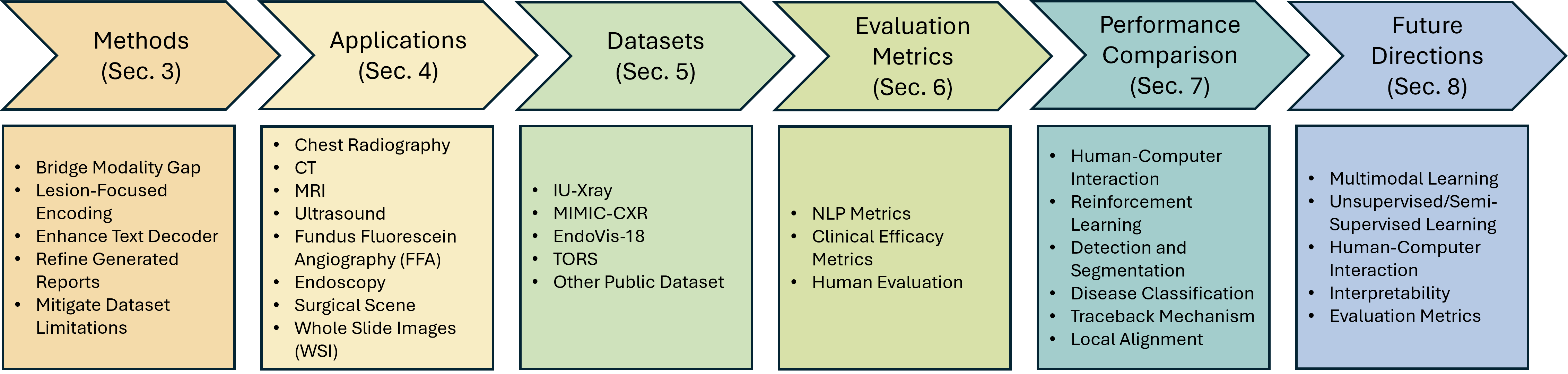}
    \caption{The content road map of this review paper. First, we present five types of solutions to address the challenges of AMRG. Next, we explore the applications of AMRG across different imaging modalities. Following this, we introduce various public datasets. Then, we outline the evaluation metrics employed to assess model performance. By comparing the performance of models on benchmark datasets, we identify six techniques that effectively enhance model performance. Finally, we discuss future research directions in the field.}
    \label{roadmap}
\end{figure}

In this review, we comprehensively examine 112 papers on AMRG based, predominantly from 2021 to 2024, and summarize various solutions proposed to address the aforementioned challenges. Our scope extends beyond radiographic report generation to include emerging applications in modalities such as MRI, CT, and ultrasound. Additionally, we present the public datasets and evaluation metrics used in this field. Through a comparative analysis of state-of-the-art (SOTA) models on benchmark radiography datasets, we identify techniques that significantly enhance evaluation metrics. Finally, we discuss potential future directions for this field. The structure of this paper is illustrated in Figure \ref{roadmap}.

\section{Problem Statement}
The objective of AMRG is to train a model that can extract meaningful features from medical images and generate descriptive text sequences that accurately describe the medical conditions depicted in the images. The primary objective function is the word-level cross-entropy loss, which measures the discrepancy between the predicted word probabilities and the actual words in the ground truth (GT) reports.

Given a medical image, the visual encoder extracts a sequence of image features $I$. The text decoder, which can be either an RNN or a Transformer model, generates a sequence of words $\{w_1, w_2, ..., w_T\}$ to describe the medical image in an autoregressive manner. At each time step $t$, the decoder generates the next word $w_t$ based on the previous words $\{w_1, w_2, ..., w_{t-1}\}$ and image features $I$. Assuming that the GT report is $\{w^*_1, w^*_2, ..., w^*_T\}$, the cross-entropy loss at each time step $t$ is given by:
\begin{equation}
    \mathcal{L}_{CE}(t) = -\log P(w^*_t|w^*_1, ..., w^*_{t-1}, I)
\end{equation}
The total loss for the entire sequence is the sum of the losses over all time steps:
\begin{equation}
\label{cross-entropy word_level}
    \mathcal{L}_{CE}=\sum^T_{t=1}\mathcal{L}_{CE}(t) = -\sum^T_{t=1}\log P(w^*_t|w^*_1, ..., w^*_{t-1}, I)
\end{equation}

\section{Methods}
\begin{figure}[htbp]
    \centering
    \includegraphics[width=0.95\textwidth]{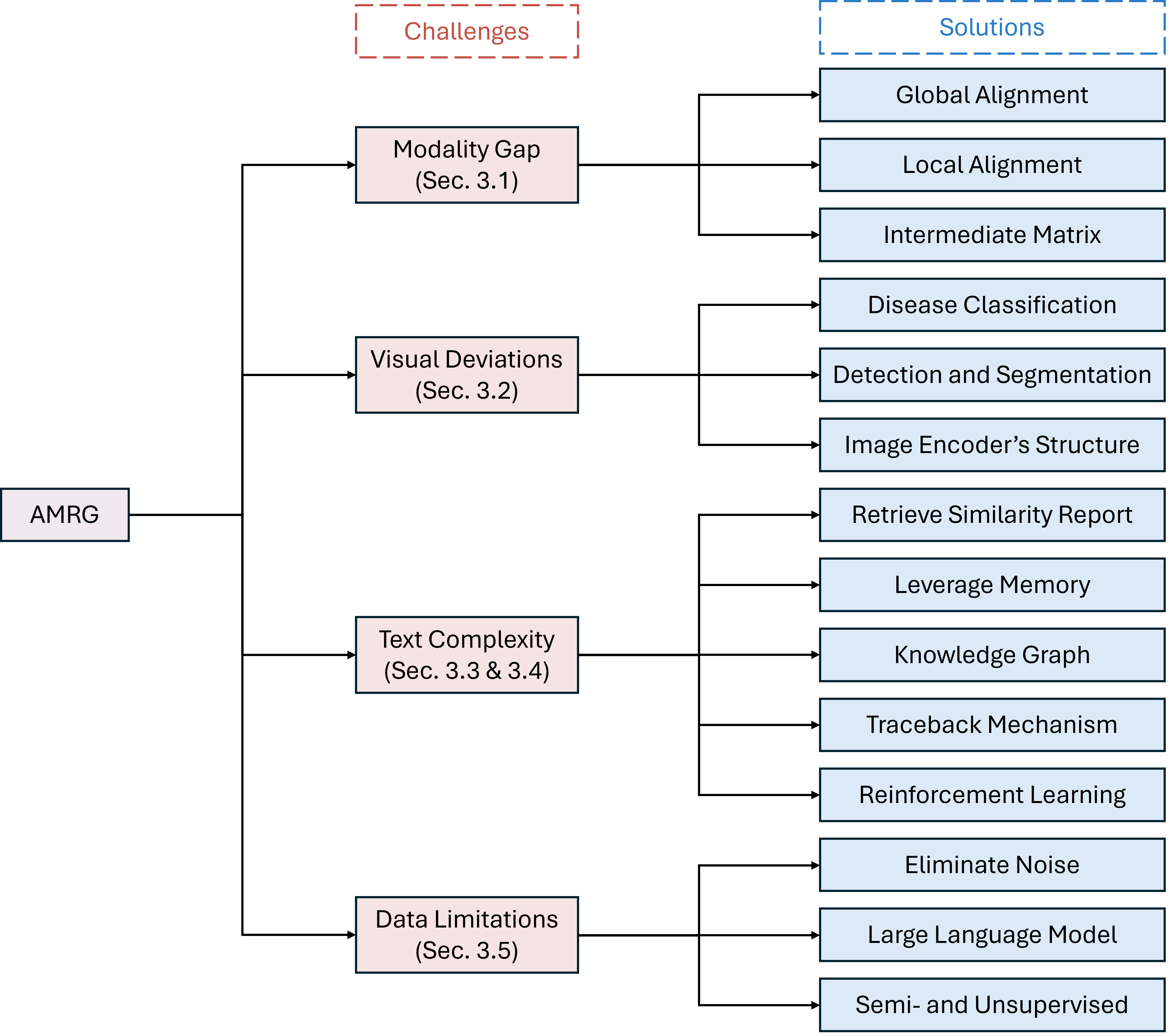}
    \caption{Four challenges in automatic medical report generation (AMRG) and their corresponding solutions.}
    \label{method}
\end{figure}

In this section, we introduce various methods designed to address the aforementioned challenges. First, we discuss techniques for bridging the gap between image-text modalities (Section \ref{Bridging the Gap Between Modalities}). Next, we present lesion-focused image encoding methods that enhance the model's ability to detect and emphasize clinically significant regions (Section \ref{Lesion-Focused Image Encoding}). We then detail approaches for enhancing the text decoder with additional information (Section \ref{Enhancing Text Decoder With Additional Information}) and refining generated reports to ensure high-quality medical outputs (Section \ref{Refining Generated Reports}). Finally, we cover strategies to mitigate dataset flaws, including methods to handle noisy and limited datasets (Section \ref{Dataset flaws}). The four main challenges and their corresponding solutions are shown in Figure \ref{method}. The following subsections delve into these solutions in detail.

\subsection{Bridging the Gap Between Modalities}
\label{Bridging the Gap Between Modalities}
\begin{figure}[htbp]
    \centering
    \includegraphics[width=1.0\textwidth]{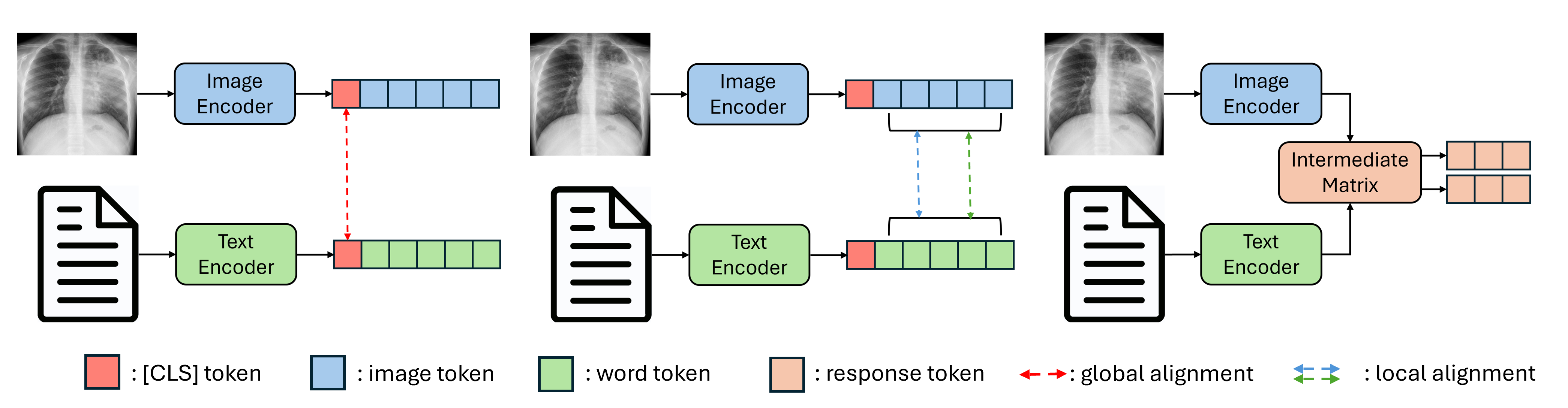}
    \caption{Flowcharts of three representative alignment methods. The left diagram illustrates global alignment, which typically uses the [CLS] token to represent the global representation of a modality. The middle diagram depicts local alignment, aligning image patches with word tokens. The right diagram shows alignment via an intermediate matrix, where a shared matrix represents the features of both modalities, ensuring they are in the same latent space.}
    \label{alignment}
\end{figure}

Bridging the gap between image and text modalities is crucial for medical report generation. This section introduces three key methods to address this challenge: (\rmnum{1}) global alignment (Section \ref{Global Alignment}), (\rmnum{2}) local alignment (Section \ref{Local Alignment}), and (\rmnum{3}) intermediate matrix alignment (Section \ref{Intermediate Matrix}). Each method offers a distinct strategy for aligning visual and textual data. Global alignment focuses on aligning entire images with entire reports to maximize mutual information and minimize discrepancies. Local alignment targets fine-grained interactions by associating specific image regions with textual elements such as sentences or words. Intermediate matrix alignment employs a shared learnable matrix to capture the alignment between visual and textual features. Figure \ref{alignment} presents simplified flowcharts of these three alignment methods. The following subsections provide detailed explanations of each method.

\subsubsection{Global Alignment}
\label{Global Alignment}
Global alignment is a method that aligns the entire image with the entire report based on InfoNCE loss \cite{oord2018representation} and triplet loss \cite{schroff2015facenet}. InfoNCE loss is well-suited for large datasets because it processes all negative pairs in a batch. In contrast, triplet loss specializes in identifying fine-grained differences by focusing on individual negative samples at a time.

The infoNCE loss function creates a joint embedding space by maximizing the cosine similarity between positive image-text pairs and minimizing it for negative pairs. This process closely aligns images and their corresponding reports. The CLIP framework \cite{radford2021learning}, which pioneers the use of InfoNCE loss for visual representation learning under natural language supervision, is particularly beneficial for report generation. This approach ensures that each medical image is effectively supervised by its paired report. Several studies have employed CLIP loss (InfoNCE loss) to successfully mitigate the modality discrepancies between radiographic images and clinical reports \cite{zhang2022contrastive, bannur2023learning, endo2021retrieval, liu2023observation, zhao2023medical}. Specifically, given a batch of N pairs of image embeddings $\{I_i\}$ and text embeddings $\{T_i\}$, the CLIP loss can be formulated as follows:
\begin{equation}
\label{InfoNCE}
    \begin{aligned}
    \mathcal{L}^{I \rightarrow T}_{IN}(I,T) &= -\frac{1}{N}\sum^N_{i=1}\log \frac{\exp(S(I_i, T_i)/\tau_1)}{\sum^N_{j=1}\exp(S(I_i, T_j)/\tau_1)}\\
    \mathcal{L}^{T \rightarrow I}_{IN}(I,T) &= -\frac{1}{N}\sum^N_{i=1}\log \frac{\exp(S(T_i, I_i)/\tau_1)}{\sum^N_{j=1}\exp(S(T_i, I_j)/\tau_1)}\\
    \mathcal{L}_{CL}(I,T) &= \frac{1}{2}(\mathcal{L}_{IN}^{I \rightarrow T}(I,T) + \mathcal{L}_{IN}^{T \rightarrow I}(I,T)),
    \end{aligned}
\end{equation}
where $\tau_1$ is a temperature parameter that scales the logits and $S(\cdot,\cdot)$ denotes cosine similarity.

However, CLIP's single-view supervision inadequately captures the intricate semantic relationships between images and text. To address this limitation, CXR-CLIP \cite{you2023cxr} employs a multi-view supervision (MVS) technique \cite{li2021supervision} that enhances training efficacy by incorporating multiple views. For instance, each chest X-ray study includes images from both the postero-anterior and lateral views (denoted as $I^1$, $I^2$), along with two text descriptions, findings, and impressions (denoted as $T^1$, $T^2$). The CLIP loss is expanded to MVS loss:
\begin{equation}
    \mathcal{L}_{MVS} = \frac{1}{4}(\mathcal{L}_{CL}(I^1,T^1)+\mathcal{L}_{CL}(I^2,T^1)+\mathcal{L}_{CL}(I^1,T^2)+\mathcal{L}_{CL}(I^2,T^2)).
\end{equation}

Furthermore, the triplet loss, another significant contrastive loss function, ensures that an anchor sample's embedding is closer to a positive sample than any negative sample by at least a predefined margin $\alpha$. This loss function has been particularly effective in the medical field, enhancing discrimination between closely resembling reports and images \cite{wang2021self,li2023unify}. Based on the paired image embeddings $I$ and text embeddings $T$ extracted by two unimodal encoders, the hardest negative samples $\tilde{I}$, $\tilde{T}$ in the batch of $N$ pairs are selected by their highest similarity to the corresponding GT modality. The triplet loss is optimized as follows:
\begin{equation}
\label{triplet}
    \mathcal{L}_{triplet} = \frac{1}{N}\sum^N_{i=1}[\alpha-S(I,T)+S(I,\tilde{T}) ]_+ +  [\alpha-S(I,T)+S(\tilde{I},T) ]_+,
\end{equation}
where $\alpha$ is the margin value and $[\cdot]_+$ represents the positive part (i.e., $\max(0,\cdot)$). It is noteworthy that combining InfoNCE loss and triplet loss can yield synergistic effects, enhancing model performance \cite{ji2021improving}.

Moreover, recognizing the limitations of using two unimodal encoders for distinguishing hard negative samples, Li et al. \cite{li2021align} recommend a multimodal encoder strategy to explore more complex modal interactions. Specifically, the image and text embeddings are jointly input into a multimodal encoder to predict whether the image and text match. Further studies have validated this approach, confirming the robustness of the multimodal encoder strategy in medical report generation \cite{li2023dynamic, jeong2024multimodal}. Additionally, exploring the use of a text decoder to generate image-text matching scores offers another avenue for optimizing the overall model beyond just the encoders \cite{wang2022automated}.

\subsubsection{Local Alignment}
\label{Local Alignment}
Although global alignment is an effective and widely adopted method, contrasting the entire image with the entire report can result in overlooking fine-grained interactions between different modalities. To address this limitation, researchers have introduced two local alignment strategies: sentence-region alignment and word-region alignment.

Sentence-region alignment matches specific image regions to corresponding sentences within a report. PhenotypeCLIP \cite{wang2023fine} employs cross-attention to generate sentence-based local textual and visual representations, replacing the global representations used in the InfoNCE loss (Equation \ref{InfoNCE}) to enhance contrastive learning. Further refining this approach, PRIOR \cite{cheng2023prior} replaces the softmax function in the cross-attention mechanism with a sigmoid function, which generates a sparser matrix and enhances computational efficiency. Moreover, PRIOR substitutes the InfoNCE loss, traditionally used in report-to-image local alignment, with a loss function based on cosine similarity and asymmetrical projection. This modification mitigates the risk of feature collapse, which results from the misclassification of positive image regions as negative.  Specifically, given a batch of $N$ pairs of image embeddings $I = \{I_1,I_2,..., I_N\}$ and report embeddings $T=\{T_1,T_2,..., T_N\}$, each image embedding $I_i$ is composed of patches $I_i = \{I_i^1,I_i^2,..., I_i^V\}$, and each report embedding $T_i$ is composed of sentences $T_i=\{T_i^1,T_i^2,..., T_i^U\}$. Here, $V$ and $U$  represent the number of image patches and sentences within a report, respectively. For each sentence $u$, the attention-based visual representation is formulated as:
\begin{equation}
    c^u_i = \sum^V_{v=1}\sigma(\frac{Q^IT_i^u \cdot K^II_i^v}{\sqrt{D}})V^II_i^v.
\end{equation}
Similarly, for each image region $v$, the attention-based textual representation is formulated as:
\begin{equation}
    c^v_i = \sum^U_{u=1}\sigma(\frac{Q^RI_i^v \cdot K^RT_i^u}{\sqrt{D}})V^RT_i^u,
\end{equation}
where $Q^I$, $K^I$, $V^I$, $Q^R$, $K^R$, $V^R$ are learnable matrices, $\sigma(\cdot)$ is the sigmoid function, and $D$ is the dimension of embeddings. The new report-to-image local alignment loss is formulated as:
\begin{equation}
    \mathcal{L}_l^{T \rightarrow I} = 
    -\frac{1}{NV}\sum^N_{i=1}\sum^V_{v=1}\frac{1}{2}
    [S(h(I^v_i),SG(c^v_i))+S(h(c^v_i),SG(I^v_i))],
\end{equation}
where $h$ is a MLP head and $SG$ denotes the stop-gradient operation.

Considering that image embeddings $I_i = \{I_i^1,I_i^2,..., I_i^V\}$ and report embeddings $T_i=\{T_i^1,T_i^2,..., T_i^U\}$ still exhibit significant differences, Liu et al. \cite{liu2024multi} introduce intermediate topics (anatomical entities) to further encode $I_i$ and $T_i$ into topic features $I^*_i = \{ I^{*1}_i,I^{*2}_i,...,I^{*M}_i\}$ and $T^*_i= \{T^{*1}_i,T^{*2}_i,...,T^{*M}_i\}$, where $M$ is the number of topics, and use weighted summation to obtain the local visual representation $c_i$, which is then utilized for local alignment:
\begin{equation}
    I^*_i = Transformer(I_i), \, T^*_i = softmax(l(T_i)) \in \mathbb{R}^M, \, c_i = \sum^M_{m=1} T^{*m}_i I^{*m}_i,
\end{equation}
where $l$ is the linear projection.

Besides sentences, words also possess varying significance within a report. For example, descriptions of abnormalities are more important than descriptions of normal findings. Therefore, researchers have proposed word-region alignment to capture more fine-grained multimodal interactions \cite{huang2021gloria, bannur2023learning, dawidowicz2023limitr, chen2023fine}. GLoRIA \cite{huang2021gloria} learns attention weights that prioritize different image regions based on their relevance to a given word and implements local contrastive learning using attention-weighted image representations. Specifically, given a pair of image embeddings $I = \{I_1, I_2,...,I_V\}$ and word embeddings $W=\{W_1, W_2,...,W_T\}$, where $V$ and $T$ represent the number of image patches and words in a report, respectively. First, the word-region similarity $s$ is calculated using the dot product, then softmax normalization is applied to obtain the attention weight $a_ {tv}$. The attention-weighted sum then forms the image representation $c_{t}$ for the word $W_{t}$:
\begin{equation}
\label{dot_product}
     s=I^TW, \quad a_{tv} = \frac{\exp(s_{tv}/\tau_2)}{\sum^V_{k=1}\exp(s_{tk}/\tau_2)}, \quad c_t = \sum^V_{v=1}a_{tv}I_v,
\end{equation}
where $s_{tv}$ denotes the similarity between the word $W_t$ and image patch $I_v$, and $\tau_2$ is a temperature parameter. Next, an aggregation function $Z$ combines the similarities between all words $W_t$ and their corresponding weighted image representations $c_t$, replacing the cosine similarity in the InfoNCE loss (Equation \ref{InfoNCE}):
\begin{equation}
\label{aggregate}
    Z(I,W) = \log (\sum^T_{t=1}\exp (S(c_t,W_t)/\tau_3))^{\tau_3},
\end{equation}
where $\tau_3$ is a temperature parameter. Dawidowicz et al. \cite{dawidowicz2023limitr} modifies this method by substituting the dot product in Equation \ref{dot_product} with element-wise multiplication and using a self-attention weighted sum to aggregate the similarities instead of a simple summation in Equation \ref{aggregate}. This modification achieves better results in various downstream tasks compared to GLoRIA.

The aforementioned methods rely on a pre-defined patch size across images. In medical image, lesions can exhibit a wide range of shapes and sizes. A fixed partition of image patches may result in incomplete or ambiguous representations of the key imaging abnormalities. Chen et al. \cite{chen2023fine} propose a method that splits an image into adaptive patches of variable sizes and aligns them with words. This method uses additional Transformer blocks and fully connected layers to predict the offset and new patch size for each adaptive patch. Then, it uniformly resamples feature points within these adaptive patches from the input image.

\subsubsection{Intermediate Matrix}
\label{Intermediate Matrix}
In addition to contrastive learning, another method to bridge the modal gap is the use of a learnable shared matrix to capture the alignment between images and texts. Chen et al. \cite{chen2022cross} propose an approach that maps visual features and textual features into a unified intermediate space. Specifically, given an embedding $I = \{I_1, I_2, ..., I_V\}$ extracted from the image and an embedding $W = \{W_1, W_2, ..., W_{t-1}\}$ extracted from the generated report, these embeddings are mapped to visual memory responses $R^I = \{R_1, R_2, ..., R_V\}$ and textual memory responses $ R^W = \{R_1, R_2, ..., R_{t-1}\}$. Both $R^I$ and $R^W$ are derived from a shared memory matrix $M$. Subsequently, $R^I$ and $R^W$ are fed into the text decoder to generate the next word at the time step $t$. The effectiveness of this method has been verified by studies from Qin et al. \cite{qin2022reinforced} and You et al. \cite{you2022jpg}.

Wang et al. \cite{wang2022cross} further improve this mapping method. They concatenate the embeddings of image-report pairs with the same disease label and apply K-means clustering to initialize the memory matrix. Moreover, they integrate both the image and text embeddings $I$ and $W$, along with visual and textual responses $R^I$ and $R^W$, and feed them into the decoder to enrich the generated content. Additionally, they incorporate triplet contrastive loss (Equation \ref{triplet}) into the optimization process to enhance the alignment of the visual and textual memory responses via explicit supervision signals. Similarly, Li et al. \cite{li2023unify} also employ triplet loss to align visual and textual features post-mapping and utilize a dual-gate mechanism to more intricately fuse visual and textual features both before and after mapping. 

\subsection{Lesion-Focused Image Encoding}
\label{Lesion-Focused Image Encoding}
\begin{figure}[htbp]
    \centering
    \includegraphics[width=1.0\textwidth]{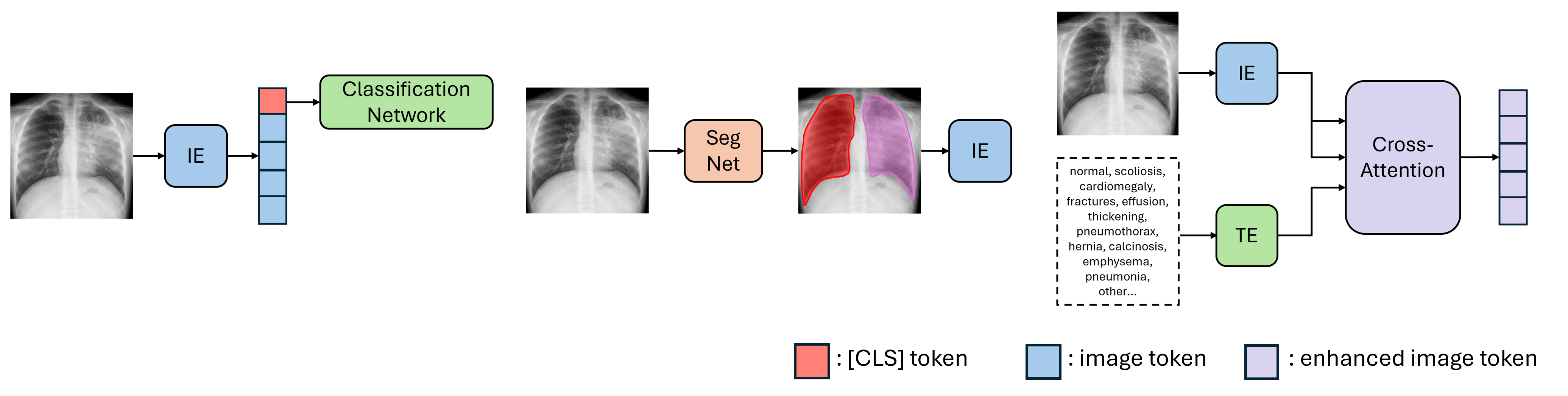}
    \caption{Flowcharts of three representative methods for enhancing image encoding: The left diagram shows that the image features extracted by the image encoder (IE) are used for disease classification, typically using only the [CLS] token instead of all image tokens. The middle diagram illustrates that the image is first processed through a pre-trained segmentation network (Seg Net) to segment meaningful areas (such as the left and right lungs), and only these areas are then input into the image encoder to eliminate background interference. The right diagram demonstrates that with cross-attention, the image features are used as keys and values, while the disease tags are used as queries. This method encourages the model to focus on image areas related to disease tags. TE represents text encoder.}
    \label{encoder}
\end{figure}

This section outlines three methods for enhancing the image encoder to focus on lesion areas and generate discriminative image representations: (\rmnum{1}) using a classification task for joint learning (Section \ref{Disease Classification}), (\rmnum{2}) employing pre-trained detection and segmentation networks as auxiliaries (Section \ref{Detection and Segmentation}), and (\rmnum{3}) modifying the internal structure of the image encoder (Section \ref{Internal Structure of Image Encoder}). The simplified flowcharts of these three methods are shown in Figure \ref{encoder}. The following subsections detail these approaches, elucidating how each method refines the encoder's capacity to identify and concentrate on clinically significant regions.

\subsubsection{Disease Classification}
\label{Disease Classification}
Adopting the features extracted by the image encoder for multi-label disease classification is an effective joint learning strategy to adapt the encoder for the report generation task \cite{you2021aligntransformer, hou2023organ, yang2023radiology, liu2023observation,jing2017automatic, wang2022medical, wang2022inclusive, wang2022automated, zhang2020radiology, wen2020symptom}. This strategy enables the image encoder to focus on regions where diseases are likely to occur and refine its ability to extract discriminative features, which helps decode accurate text. As a result, it enhances the encoder's sensitivity to medically relevant areas and clinically significant details indicative of various diseases.

Nevertheless, due to the distinct operational mechanisms of CNNs and Transformers, their implementation approaches exhibit slight variations.  For CNN-based image encoders, all features from the last convolutional layer are used for classification \cite{you2021aligntransformer, hou2023organ, jing2017automatic}, often coupled with average pooling to achieve a global representation \cite{yang2023radiology, liu2023observation}. This approach may result in image features that contain only high-level information for classification, while losing the low-level information necessary for generating descriptive text. Conversely, Transformer-based image encoders utilize an independent [CLS] token to extract global features by interacting with other image patch tokens \cite{wang2022medical, wang2022inclusive, wang2022automated}. Only the [CLS] token is used for classification, which prevents the image tokens from being encoded too abstractly. Additionally, feeding the classification results into the text decoder improves the quality of the generated reports. \cite{you2021aligntransformer, hou2023organ, liu2023observation, wang2022medical, wang2022inclusive, jin2024promptmrg, wen2020symptom}. 

A single [CLS] token may not accurately cover all diseases, similar to how a general practitioner’s diagnosis may not be as precise as that of a specialist. METransformer \cite{wang2023metransformer} addresses this by concatenating multiple expert tokens in the image encoder and using orthogonal loss to minimize overlap among these tokens, thereby encouraging them to capture complementary information. The model generates a report based on each expert token and selects the best report through a voting strategy.

\subsubsection{Detection and Segmentation}
\label{Detection and Segmentation}
In addition to using classification tasks to guide the image encoder towards clinically relevant areas, researchers have proposed using pre-trained segmentation or detection networks to explicitly assist the encoder in targeting anatomical regions \cite{zhao2023medical, zhang2024sam, wang2023self, tanida2023interactive}. One approach employs SAM \cite{kirillov2023segment} to segment meaningful anatomical regions (e.g., the left and right lungs) from chest X-ray images before inputting them into the image encoder \cite{zhao2023medical}. This method eliminates background interference, thereby enhancing the encoder's focus on relevant regions. Other researchers combine global features extracted by the image encoder with regional features from the detection or segmentation network, inputting both into the decoder to enrich the information it receives \cite{zhang2024sam, wang2023self}. In addition, because lesions typically occupy a small portion of medical images, Tanida et al. \cite{tanida2023interactive} propose a framework that compels the model to focus on the critical regions. This framework involves cropping multiple anatomical regions from the input image using a detection network, followed by multiple binary classification networks to evaluate whether each region is critical for report generation. The text decoder processes only the critical regions, thereby preventing it from being overwhelmed by the numerous normal regions.

\subsubsection{Internal Structure of Image Encoder}
\label{Internal Structure of Image Encoder}
In addition to adopting a joint learning strategy and using pre-trained auxiliary networks, modifying the internal structure of the image encoder can also enhance its focus on lesion areas. Two effective and widely used methods are cross-attention and high-order attention.

The cross-attention mechanism \cite{vaswani2017attention} assigns weights to image regions based on their relevance to disease tags, thereby emphasizing the features of regions containing lesions. This method does not require disease annotations for each image, but rather a set of all disease tags \cite{cao2023mmtn, you2021aligntransformer, liu2021exploring}. Specifically, the disease tag set is used as the query, and the image is used as the key and value. The dot product in the cross-attention mechanism can select disease tags related to the image content and enhance the features of regions containing these diseases.

Recently, several studies have attempted to replace traditional first-order attention with X-linear attention \cite{pan2020x} in Transformer-based image encoders \cite{wang2022medical, wang2023metransformer, xu2023hybrid, wang2023mvco}. X-linear attention captures complex high-order interactions within medical images, leading to a more nuanced and comprehensive understanding of the images and more accurate localization of abnormalities. In detail, given a query $Q \in \mathbb{R}^{D_q}$, a set of keys $K=\{k_i\}^N_{i=1}$ and a set of values $V=\{v_i\}^N_{i=1}$, where $k_i \in \mathbb{R}^{D_k} $ and  $v_i \in \mathbb{R}^{D_v} $, low-rank bilinear pooling \cite{kim2016hadamard} is performed to obtain the joint bilinear query-key $B_k$ and query-value $B_v$:
\begin{equation}
    B^k_i = \sigma (W_k k_i) \odot \sigma (W_q^k Q), \quad B^v_i = \sigma (W_v v_i) \odot \sigma (W_q^v Q),
\end{equation}
where $W_k \in \mathbb{R}^{D_B \times D_k}$,  $W_v \in \mathbb{R}^{D_B \times D_v}$, and $W_q^k, W_q^v \in \mathbb{R}^{D_B \times D_q}$ are learnable matrices, $\sigma$ denotes ReLU unit, and $\odot$ represents element-wise multiplication. Then, the spatial attention $\beta_i^s$ and channel-wise attention $\beta^c$ are computed as follows:
\begin{equation}
\begin{aligned}
     B^{'k}_i &= \sigma (W^k_B B_i^k), \quad \beta^s_i = softmax(W_s B^{'k}_i)\\
    \bar{B} &= \frac{1}{N}\sum^N_{i=1} B^{'k}_i, \quad \beta^c = sigmoid (W_c \bar{B}),
\end{aligned}
\end{equation}
where $W_B^k \in \mathbb{R} ^{D_c \times D_B}$, $W_s \in \mathbb{R}^{1 \times D_c}$, and $W_c \in \mathbb{R}^{D_B \times D_c}$ are learnable matrices. Finally, the output of the X-linear attention mechanism is given by:
\begin{equation}
    \hat{v} = F_{X-linear}(K,V,Q) = \beta^c \odot \sum^N_{i=1} \beta^s_i B^v_i
\end{equation}

\subsection{Enhancing Text Decoder With Supplementary Information}
\label{Enhancing Text Decoder With Additional Information}
\begin{figure}[htbp]
    \centering
    \includegraphics[width=1.0\textwidth]{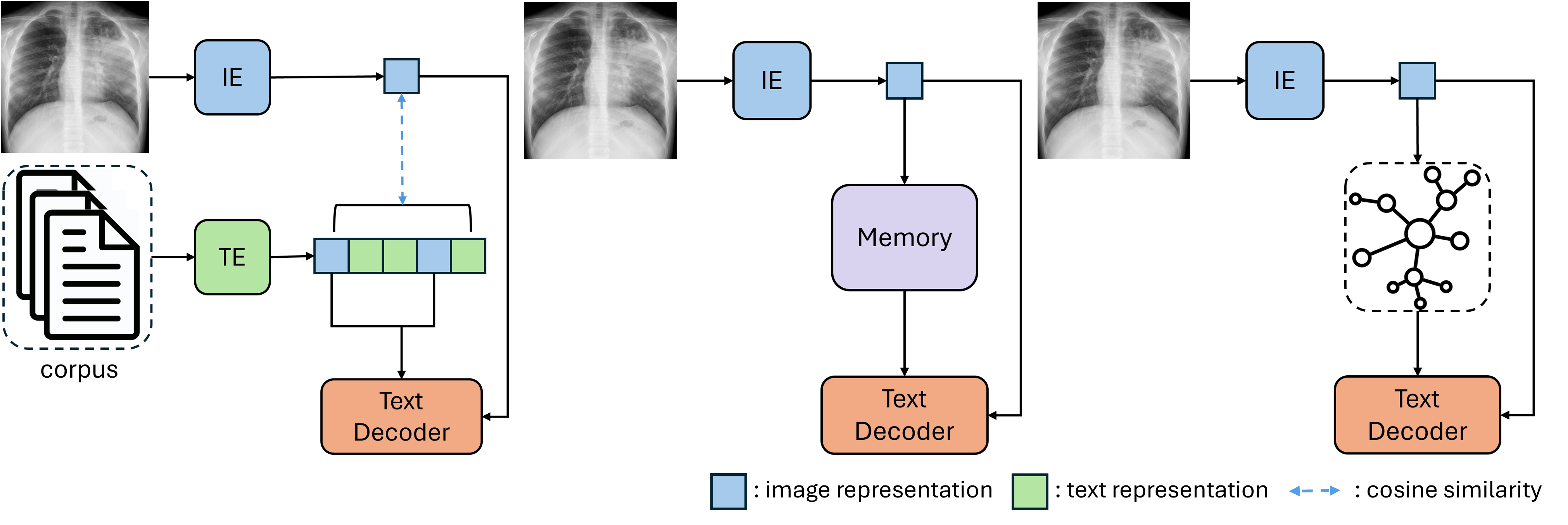}
    \caption{Flowcharts of three representative methods for augmenting the text decoder with supplementary information. The left diagram shows a retrieval-based approach. Reports similar to the input image are found from the corpus (consisting of training reports) based on cosine similarity and are input into the text decoder as reference information. The middle diagram illustrates replacing the corpus with a learnable memory to avoid leakage of training data. The right diagram demonstrates replacing the memory with a knowledge graph to store the clinical information used to generate the report in a structured manner. IE and TE represent image encoder and text encoder, respectively.}
    \label{decoder}
\end{figure}

This section presents three approaches for augmenting the text decoder with supplementary information: (\rmnum{1}) retrieving similarity reports (Section \ref{Retrieve Similarity Reports}), (\rmnum{2}) leveraging memory (Section \ref{Memory}), and (\rmnum{3}) integrating knowledge graphs (Section \ref{Knowledge Graph}). Each approach addresses specific challenges, such as ensuring clinical consistency, alleviating privacy concerns, and building medical knowledge for better model comprehension. Figure \ref{decoder} shows the flowcharts of the three methods.

\subsubsection{Retrieve Similarity Reports}
\label{Retrieve Similarity Reports}
Given the limited diversity of diagnoses in medical reports, a large retrieval corpus can adequately cover the potential diagnoses of input images. Some researchers have proposed using a retrieval-based approach to generate new reports, with the primary advantage being the clinical consistency of the generated reports with manually written ones \cite{endo2021retrieval, jeong2024multimodal}. To elaborate, image and text encoders are trained using the CLIP method, which produces higher similarity scores for paired image-text examples and lower scores for unpaired ones. A corpus is then constructed using the reports in the training set. During inference, the model retrieves the top $K$ reports from the corpus with the highest similarity scores to the input image and combines them into the predicted report. However, since candidate selection is based on maximizing similarity scores, the predicted report is prone to repeating information.

PPKED \cite{liu2021exploring} improves the basic retrieval-based approach by modifying the retrieval process and using a text decoder to generate reports instead of merely combining retrieved candidates. The corpus is constructed using image-text pairs from the training set. During inference, the system retrieves the top $K$ images in the corpus that are most similar to the input image and uses their corresponding reports to enhance the image features. Finally, the text decoder generates the final report based on the enhanced image features, ensuring coherence and the absence of redundant content.

\subsubsection{Memory}
\label{Memory}
However, retrieving training data during inference raises concerns regarding the privacy of medical data. Some researchers have proposed a solution by employing learnable memory to replace the corpus \cite{yang2023radiology, cheng2023prior}. The memory stores features derived from the training data, rather than the training data itself, thereby mitigating the risk of data leakage. Yang et al. \cite{yang2023radiology} use cross-attention to update the memory during training and to enhance image features during inference. Cheng et al. \cite{cheng2023prior} adopt a more explicit approach to update the sentence-prototype memory. During training, the sentence prototype most similar to the input sentence is selected using cosine similarity, and the memory is updated based on the L1 loss between the prototype and the input sentence.

Furthermore, integrating memory into the text decoder is another method to enhance the quality of the generated reports \cite{liu2021auto, wang2022medical, chen2020generating, wang2023fine}. This memory records fine-grained medical knowledge and historical information from previous generation processes, which is valuable for generating lengthy texts. One approach involves using the memory matrices to augment the keys and values of the Transformer-based decoder \cite{liu2021auto, wang2022medical}. Specifically, given a key $K$ and value $V$, the memory-augment key and value are defined as $\hat{K} = [K, M_k]$ and $\hat{V} = [V, M_v]$, where $M_k$ and $M_v$ are learnable matrices, and $[\cdot,\cdot]$ denotes concatenation. Another approach utilizes a gate mechanism to update the memory and map it to the scale and offset parameters in layer normalization, thereby injecting the memory into the decoder \cite{chen2020generating, wang2023fine}.

\subsubsection{Knowledge Graph}
\label{Knowledge Graph}
A more structured memory, in the form of a medical knowledge graph, can group diseases according to organs or body parts. This is because abnormalities in the same body part often exhibit strong correlations and common features. Initially, a medical graph is designed based on prior knowledge from chest findings to cover common abnormalities and their relationships \cite{zhang2020radiology}. In this graph, disease keywords serve as nodes ($V$) and their relationships as edges ($E$), denoted as $G=\{V,E\}$. Graph convolution \cite{kipf2016semi} is used to propagate information within the graph, thereby enhancing the model's capacity to comprehend medical knowledge. This pre-constructed graph has been adopted by several studies, which further distill the knowledge graph during the decoding stage to enrich the information received by the decoder. \cite{zhang2023semi, huang2023kiut, liu2021exploring}.

However, a fixed graph may not contain all the necessary knowledge about the input image, thereby limiting its effectiveness. Li et al. \cite{li2023dynamic} design a dynamic graph that uses the nodes from the pre-constructed graph \cite{zhang2020radiology} as initial nodes and models relationships with an adjacency matrix. In this matrix, 1 represents a connection between two nodes, while 0 indicates no relationship. The dynamic update process is as follows: during training, the model retrieves the top three most similar reports from a corpus for each input image. Then, RadGraph \cite{jain2021radgraph} is applied to extract specific knowledge triplets from these reports, formatted as $\{subject \; entity,\; relation, \allowbreak \; object \; entity\}$. If only the subject or object entity is present in the graph, the other entity in the triplet is added as an additional node, and their relation is set to 1 in the adjacency matrix, indicating a link between the two nodes.

In addition to learning the relationships between entities, MGSK \cite{yang2022knowledge} uses more explicit and accurate relationships that are manually annotated. The model comprises two graphs: a general graph and a specific graph. The general graph is independent of the input image and is manually constructed by radiologists from 500 radiology reports in the MIMIC-CXR dataset \cite{jain2021radgraph}. The general knowledge is stored in the triplet, $\{subject \; entity,\; relation,\allowbreak \; object \; entity\}$, where relations include ‘suggestive of,’ ‘modify,’ and ‘located at’. The specific knowledge is retrieved from the corpus by finding the top ten images most similar to the input image and extracting triplets from their corresponding reports using RadGraph. The general and specific knowledge graphs are fused with image features to enrich the input of the text decoder.

\subsection{Refining Generated Reports}
\label{Refining Generated Reports}
This section outlines methods designed to refine the accuracy and semantic coherence of generated medical reports. In particular, these techniques are designed to guarantee that the generated content accurately reflects essential medical insights that are critical for clinical reliability. Two pivotal approaches will be discussed: (\rmnum{1}) the traceback mechanism (Section \ref{Traceback Mechanism}) , which evaluates semantic fidelity, and (\rmnum{2}) reinforcement learning (Section \ref{Reinforcement Learning}), which correlates training goals with evaluative metrics.

\subsubsection{Traceback Mechanism}
\label{Traceback Mechanism}
\begin{figure}[htbp]
    \centering
    \includegraphics[width=0.65\textwidth]{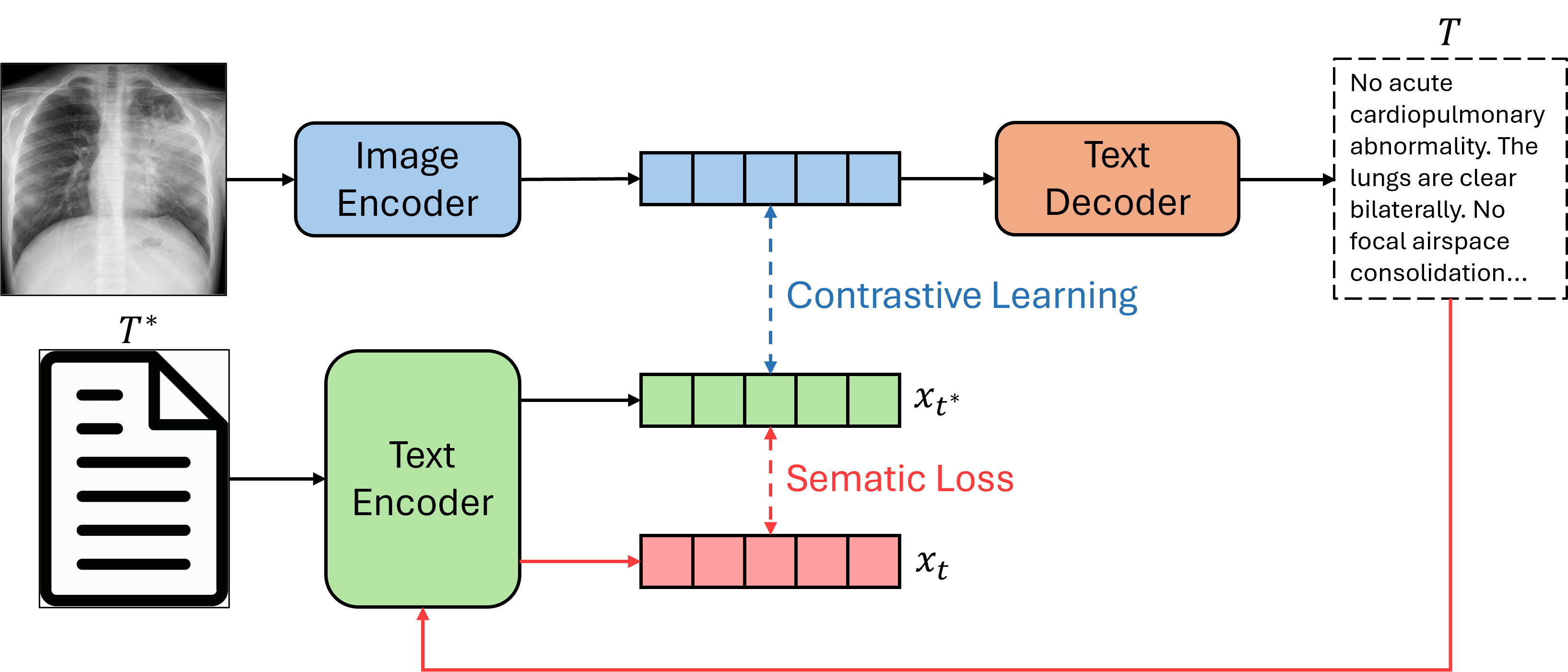}
    \caption{The GT report $T^*$ is input into the text encoder to obtain semantic features $x_{t^*}$ (text representation). The visual representation extracted by the image encoder is then fed into the text decoder after contrastive learning, producing the generated report $T$. The traceback mechanism begins by inputting this generated report $T$ back into the text encoder to extract the semantic features $x_t$. The difference between $x_{t^*}$ and $x_t$, termed the semantic loss, serves as the objective function of the traceback mechanism. This mechanism aims to reduce the discrepancy between the generated report $T$ and the GT report $T^*$ at the feature level.}
    \label{traceback}
\end{figure}

Most medical report generation methods construct loss functions that evaluate the discrepancy between generated and GT reports at the word level (for further details, please refer to Equation \ref{cross-entropy word_level}). Consequently, models tend to predict frequently observed words in order to achieve a high overlap rate \cite{holtzman2019curious}, which may result in the generation of clinically flawed reports. A high-quality generated report should also be semantically similar to the GT. To achieve this, some researchers propose a traceback mechanism to control the semantic validity of generated content through self-assessment \cite{wang2021self, li2022self, ye2024dual, chen2023fine}.  This approach involves inputting the generated report $T$ into a text encoder to extract semantic features $x_{t}$, and optimizing the model to ensure that $x_{t}$ is similar to the semantic features $x_{t^*}$ of the GT report $T^*$. The process of the traceback mechanism is shown in Figure \ref{traceback}.

A variety of techniques exist for measuring semantic loss, including calculating the L2-norm distance \cite{wang2021self} and the cosine similarity \cite{li2022self} between $x_{t}$ and $x_{t^*}$, as well as using a classifier to ensure that the disease classifications based on $x_{t}$ and $x_{t^*}$ are the consistent \cite{ye2024dual}. Classifying diseases based on the semantic features is a preferable approach because it encourages the model to correct generated words that influence disease classification, which is the most critical aspect of medical reports. A more complex approach involves having the model synthesize a medical image $I$ based on the generated report $T$ and comparing $I$ with the input image $I^*$ \cite{chen2023fine}. Moreover, to reduce significant gradient fluctuations during the initial training stages, it is advantageous to assign a smaller weight to the traceback semantic loss at the beginning \cite{ye2024dual}.

\subsubsection{Reinforcement Learning}
\label{Reinforcement Learning}
\begin{figure}[htbp]
    \centering
    \includegraphics[width=1.0\textwidth]{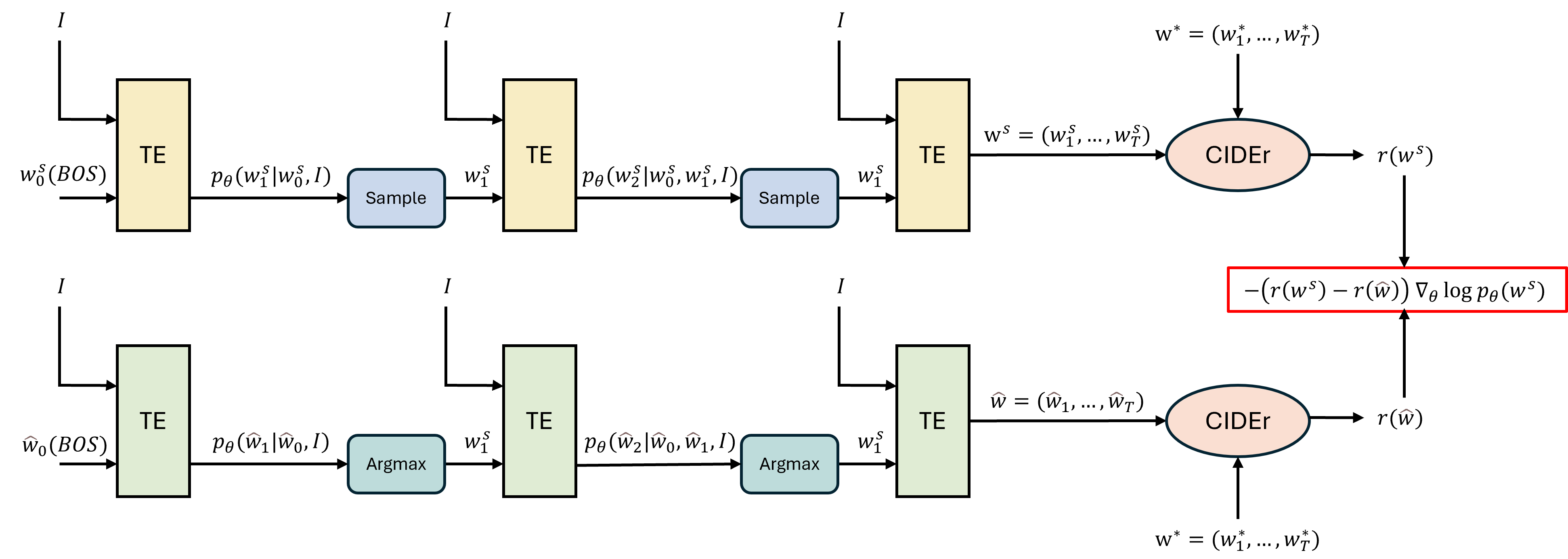}
    \caption{Implementation of reinforcement learning involves several key elements. \( I \) represents the image representation, while BOS is a special token denoting the beginning of the sequence. \( w^s \) is the sequence obtained by sampling, \( \hat{w} \) is the sequence obtained by the argmax operation, and \( w^* \) is the GT sequence. By calculating the CIDEr scores for \( w^s \) and \( w^* \), as well as for \( \hat{w} \) and \( w^* \), the rewards \( r(w^s) \) and \( r(\hat{w}) \) are obtained. The gradient of the objective function, highlighted in the red box, is then computed based on these rewards. TE represents text decoder.}
    \label{reinforcement}
\end{figure}

In contrast to indirect semantic loss, some researchers use reinforcement learning with NLP metrics as rewards to align training goals with final evaluation criteria \cite{lin2023sgt++, wang2021self, wang2022medical, cornia2020meshed, qin2022reinforced, liu2024multi}. Specifically, the text decoder is treated as ``agent'' that interacts with an external ``environment'' (visual and textual features). The network parameters, $\theta$, define a ``policy'' $p_\theta$, that results in an ``action'' (the prediction of the next word). The CIDEr score is used as a reward $r$, which is calculated by comparing the generated sequence to the corresponding GT sequence. The objective of training is to minimize the negative expected reward:
\begin{equation}
    L(\theta) = - \mathbb{E}_{w^s \sim p_\theta}[r(w^s)],
\end{equation}
where $w^s = (w^s_1, ..., w^s_T)$ and $w^s_t$ represents the word sampled from the model at the time step $t$. The expected gradient of the non-differentiable reward function can be approximated using a Monte-Carlo sample $w^s = (w^s_1, ..., w^s_T)$ from $p_\theta$:
\begin{equation}
    \nabla_\theta L(\theta) \approx - (r(w^s)-r(\hat{w}))\nabla_\theta \log p_\theta(w^s),
\end{equation}
where $r(\hat{w})$ is the reward obtained by the current
model under the inference algorithm at test time, and $\hat{w}$ is generated by greedy decoding:
\begin{equation}
    \hat{w}_t = \arg \max_{w_t} p(w_t|\hat{w}_0,..., \hat{w}_{t-1}, I),
\end{equation}
where $\hat{w}_0,..., \hat{w}_{t-1}$ are the previous generated words and $I$ is the image representation. As a result, samples $w^s$ from the model that yield a higher reward than $\hat{w}$ increase their probability during the learning process, whereas samples resulting in a lower reward are suppressed. The process of implementing reinforcement learning is shown in Figure \ref{reinforcement}.

However, using only one NLP metric (e.g. CIDEr) as a reward may lead to partial optimization rather than overall optimization, as long text generation tasks cannot rely on a single metric to evaluate performance. Xu et al. \cite{xu2023hybrid} test seven NLP metrics with different combinations as rewards and find that using BLEU-4, METEOR, and CIDEr together achieves the best results. Miura et al. \cite{miura2020improving} design the factual completeness and consistency rewards that are more suitable for medical reports. These special rewards serve to ascertain whether the generated report and the GT report contain the same anatomical entities and whether the sentences containing these entities contradict the corresponding sentences in the GT report.

\subsection{Dataset Limitations}
\label{Dataset flaws}
Medical image-text paired datasets are constrained by two fundamental limitations: noise and limited size. Noise arises from temporal information and false negatives, which can distort the training data and lead to inaccurate model predictions. Temporal information noise occurs when reports reference earlier images, which introduces inconsistencies that models struggle to interpret correctly. False negatives, on the other hand, arise in contrastive learning when similar reports are incorrectly treated as negative samples, confusing the model. Additionally, the limited size of these medical datasets presents another challenge, as insufficient data hinders the model's capacity to generalize and perform effectively. The following subsubsections present innovative methods to address these issues: (\rmnum{1}) removing temporal noise and mitigating false negatives in contrastive learning (Section \ref{Eliminating Noise}), (\rmnum{2}) using LLM to improve model performance on limited datasets (Section \ref{Large Language Model}), and (\rmnum{3}) expanding training datasets through semi-supervised and unsupervised learning (Section \ref{Semi-Supervised and Unsupervised Learning}).

\subsubsection{Eliminating Noise}
\label{Eliminating Noise}
Although medical image-report datasets authorized by professionals are generally more accurate than general image-text datasets in terms of annotation,   they still exhibit inherent noise, including temporal information and false negatives. Temporal information noise can be attributed to reports that reference earlier images. For instance, in a report, ``\textbf{Comparison made to prior study}, there is \textbf{again} seen moderate congestive heart failure with \textbf{increased} vascular cephalization, \textbf{stable}. There are large bilateral pleural effusions \textbf{but decreased since previous}'', the bolded words relate to earlier images, yet the current image paired with the report does not contain this comparative information. Such references to previous data introduce temporal noise, which may lead trained models to generate hallucinations about non-existent priors. False negatives occur in contrastive learning when negative or unpaired texts (reports from other patients) describe identical symptoms as the paired reports. Simply treating the other reports as negative samples introduces noise into the supervision process, thereby confusing the model.

To eliminate temporal noise, Ramesh et al. \cite{ramesh2022improving} propose two approaches: rewriting medical reports using GPT-3 and designing a token classifier to delete word tokens associated with prior studies. The latter method is more accurate and cost-effective. In contrast to eliminating temporal information, some methods effectively integrate it into the learning process \cite{bannur2023learning, hou2023recap}. For example, the BioViL-T framework \cite{bannur2023learning} assumes that a patient has a current image $I_c$, a current report $T_c$, and a previous image $I_p$. The images $I_c$ and $I_p$ are processed through a CNN to generate features $P_c$ and $P_p$, respectively. These features are then fed into Transformer blocks to extract difference features $P_d$. Subsequently, $P_c$ and $P_d$ are input together into the text decoder. If the model's input includes only the current image $I_c$, $P_d$ is substituted with a learnable feature $P_m$.

To address false negatives within contrastive learning, ALBEF \cite{li2021align} utilizes momentum distillation to mitigate the impact of such noise in general image captioning datasets. The momentum encoder, functioning as a teacher, produces a stable set of pseudo-labels for the input image. These pseudo-labels serve as training targets for the student encoder to account for the potential positives in the negative pairs. Research has shown that this momentum distillation can be seamlessly adapted to medical datasets \cite{li2023dynamic, jeong2024multimodal}. Furthermore, MedCLIP \cite{wang2022medclip} decouples medical image-text pairs and employs semantic similarity to create pseudo-labels. This approach entails extracting entities from the reports as textual labels and utilizing disease labels as image labels. The cosine similarity between these image and text labels serves as pseudo-labels in contrastive learning.

\subsubsection{Large Language Model}
\label{Large Language Model}
To address the issue of the limited size of medical image-report datasets, a feasible approach is to utilize a pre-trained large language model (LLM) as a text decoder. This method leverages the LLM's robust language generation and zero-shot transfer capabilities, thereby reducing the number of parameters that need to be trained from scratch. Li et al. \cite{li2023blip} propose a trainable mapping module, Q-Former, to bridge the gap between a frozen image encoder and a frozen LLM. Specifically, they employ a pre-trained Vision Transformer (ViT) as an image encoder to extract image features, which are then mapped to the text feature space by Q-Former. Subsequently, the frozen LLM serves as a text decoder to generate reports. MSMedCap \cite{zhang2024sam} demonstrates that Q-Former is effective for handling limited medical image-report data.

Nevertheless, utilising a frozen LLM as a text decoder concurrently with a frozen image encoder is not the optimal approach. Research indicates that fine-tuning both the image encoder and the mapping module when the LLM decoder is frozen can enhance the quality of the generated reports \cite{wang2023r2gengpt}. Additionally, freezing the LLM may not be the most effective strategy, as an LLM pre-trained on general data may not be suitable for the medical domain. It is therefore generally recommended to fine-tune the LLM on task-specific data. However, it should be noted that fine-tuning an LLM requires a substantial amount of data, so directly fine-tuning an LLM with limited medical data may result in suboptimal performance.

Liu et al. \cite{liu2024bootstrapping} propose a coarse-to-fine decoding strategy to fine-tune an LLM on limited medical datasets in a bootstrapping manner. They initially employ MiniGPT-4 \cite{zhu2023minigpt} to generate a coarse report and then use the coarse report as a prompt, along with the image features, to input into the decoder of MiniGPT-4 again to generate the final refined report. Additionally, pseudo self-attention \cite{ziegler2019encoder} is another method for fine-tuning an LLM on medical data \cite{tanida2023interactive, alfarghaly2021automated}. This approach introduces new parameters solely within the self-attention block, while other parameters of the Transformer are initialized with pre-trained values.  Given an image feature $X$ and a hidden state $Y$, the pseudo self-attention is formulated as follows:
\begin{equation}
    PSA(X,Y) = softmax((YW_q)
\begin{bmatrix}
    XU_k \\
    YW_k
\end{bmatrix}^T)\begin{bmatrix}
    XU_v \\
    YW_v
\end{bmatrix},
\end{equation}
where $U_k$, $U_v$ are new parameters, and $W_q$, $W_k$, $W_v$ are parameters from pre-trained model. Fine-tuning an LLM with pseudo self-attention results in minimal changes to the pre-trained LLM’s parameters, thereby maintaining its text generation capabilities \cite{ziegler2019encoder}.

\subsubsection{Semi-Supervised and Unsupervised Learning}
\label{Semi-Supervised and Unsupervised Learning}
To address the limited size of medical image-text paired datasets, some researchers have explored semi-supervised and unsupervised learning methods to expand the training dataset. The RAMT model \cite{zhang2023semi} employs a student-teacher network to train in a semi-supervised manner using 25\% paired data and 75\% unpaired image data. The student and teacher networks share the same structure but have distinct parameters. During the training phase, the teacher network parameters are updated by the exponential moving average (EMA) of the student network parameters. Different noises are added to the input images, which are then fed into the student and teacher networks, respectively. The output of the teacher network serves as supervision for the student network.

However, the semi-supervised method still requires some images with corresponding reports. KGAE \cite{liu2021auto} addresses this limitation by utilizing unsupervised learning. The model employs a pre-constructed knowledge graph $G$ as a shared latent space to bridge the gap between image and text representations. Given an input image $I$ and an input report $R$, the graph $G$ maps them into the same latent space, $G_I$ and $G_R$. For unsupervised learning, the image encoder and text decoder are trained separately. To train the image encoder, $G_I$ and $G_R$ jointly implement disease classification, ensuring they form a common latent space. To train the text decoder, the report $R$ is reconstructed from $G_R$, following the process $R \rightarrow G_R \rightarrow R$. Additionally, KGAE can be applied in semi-supervised and supervised settings. In these settings, as well as for inference, the pipeline follows $I \rightarrow G_I \rightarrow R$.

In a more recent study, Hirsch et al. \cite{hirsch2024medcycle} refine the unsupervised method, achieving higher accuracy than KGAE. The method employs cycle consistency to ensure that cross-modal mappings retain information. Cross-modal mappings include image-to-report ($I2R$) and report-to-image ($R2I$). The objective of cycle consistency is to minimise the differences between $z_i$ and $R2I(I2R(z_i))$, as well as $z_r$ and $I2R(R2I(z_r))$, where $z_i$ and $z_r$ represent image and text representations, respectively. To ensure that the image encoder extracts relevant semantic information (e.g., diseases and organs), they employ contrastive learning to align the image representations (output by the image encoder) with the text representations of pseudo-reports. These pseudo-reports are constructed based on the disease labels of images. To train the text decoder, they also adopt a report reconstruction task and add adversarial learning to ensure that the text decoder receives similar features during training and inference.

\section{Applications}
In this section, we introduce the applications of AMRG across various medical imaging modalities, including chest radiography, computed tomography (CT), magnetic resonance imaging (MRI), ultrasound, ophthalmic imaging, endoscopy, surgical scene, and pathological imaging. These modalities are crucial for diagnosing a broad spectrum of medical conditions.

\textbf{Chest Radiography: }In recent years, a significant amount of research focused on chest radiography report generation (refer to Table \ref{tab:results}). This is largely due to the availability of large and publicly accessible datasets such as MIMIC-CXR \cite{johnson2019mimic} and IU X-ray \cite{demner2016preparing}, which contain extensive collections of annotated images and paired reports. The availability of these datasets enables the models to effectively learn the intricate relationships between visual features and textual descriptions. 

The application of AMRG in radiography offers several significant benefits. First, it can significantly reduce the radiologist's workload by automating the initial draft of the report, allowing them to focus on more complex and nuanced cases \cite{alfarghaly2021automated, wu2023multimodal, aksoy2023radiology, dyer2021diagnosis, sloan2024automated}. Second, these models can improve diagnostic consistency and reduce inter-observer variability by applying standardized criteria and guidelines in the report generation process \cite{yu2023evaluating}. Third, in regions with limited access to experienced radiologists, AMRG models can provide essential diagnostic support, ensuring timely and accurate medical care for patients \cite{mun2021artificial}. Finally, these models can support large-scale screening programs by rapidly processing and generating reports for large volumes of X-ray images, thereby facilitating the early detection of diseases.

\textbf{3D Imaging: }CT and MRI provide detailed, three-dimensional (3D) views of the human body, playing a pivotal role in diagnosing a wide range of conditions, including neurological disorders and abdominal diseases. Recent studies have explored AMRG for these imaging modalities \cite{han2021unifying,chen2024mapping,zhang2023novel}, but these studies often treat 3D images as a set of 2D slices, overlooking the inherent stereoscopic structural information, an issue that future research should address.

\textbf{Ultrasound: }Ultrasound is widely used due to its real-time imaging capability and safety profile. Recently, AMRG has been explored for ultrasound applications \cite{li2024ultrasound}. enabling real-time report generation and assisting clinicians in making immediate decisions, especially in emergency and point-of-care settings. However, the low image quality and operator-dependent nature of ultrasound image acquisition still affect the quality of generated reports.

\textbf{Ophthalmic Imaging: }In ophthalmology, AMRG applications in fundus fluorescein angiography (FFA)\cite{li2022cross} and fundus images \cite{wen2020symptom} aid in diagnosing critical eye diseases such as diabetic retinopathy. Li et al. \cite{li2022cross}, proposed the CGT model for ophthalmic report generation. Their approach involves an information extraction scheme that converts unstructured medical reports into a structured format, constructing clinical graphs. These graphs encapsulate prior medical knowledge, which is then distilled into sub-graphs and integrated with visual features to enhance report generation. By employing a combination of cross-entropy and triples loss, they optimize the report generation model, achieving SOTA results on the FFA-IR benchmark dataset \cite{li2021ffa}. 

\textbf{Endoscopy: }For endoscopic imaging, AMRG aids in diagnosing various complications, such as gastrointestinal diseases and cancers. Cao et al. \cite{cao2023mmtn} combined disease tags with cross-attention and introduced memory augmentation in the image encoder to improve the model's sensitivity to lesion areas. Their model achieved competitive results with SOTA models on the gastrointestinal endoscope image dataset, which is a private dataset contains white light images and their Chinese reports from the department of gastroenterology.

\textbf{Surgical Scene: }In surgical imaging, AMRG helps create operative reports by documenting surgical steps. This alleviates surgeons' workloads and allows them to focus more on the operations. Lin et al. \cite{lin2023sgt++} proposed the SGT++ model to effectively models interactions between surgical instruments and tissues. Their method involves homogenizing heterogeneous scene graphs to learn explicit, structured, and detailed semantic relationships via an attention-induced graph Transformer. Additionally, it incorporates implicit relational attention to integrate prior knowledge of interactions. 

\textbf{Pathological Imaging: }Pathological imaging involves examining high-resolution images of tissue sections, requiring meticulous analysis. Chen et al. \cite{chen2024wsicaptionmultipleinstancegeneration} recently introduced the MI-Gen model to produce pathology reports for gigapixel whole slide images (WSIs). Furthermore, they created the largest WSI-text dataset, PathText, which contains nearly 10,000 high-quality WSI-text pairs. 

\textbf{Unified Model: }Unlike the above models specialized for one modality, Google recently introduced a groundbreaking model, Med-PaLM M \cite{tu2024towards}, which encodes and interprets various biomedical data modalities using the same model weights. This model can process multiple data modalities, including clinical language, genomics, and imaging (e.g., radiography, mammography, dermatology, and pathology). To support these developments, they curated MultiMedBench \cite{tu2024towards}, a benchmark comprising 14 tasks such as AMRG, report summarization, medical question answering, visual question answering, and medical image classification. Med-PaLM M achieved performance competitive with or surpassing specialist models on all MultiMedBench tasks. This innovation marks a significant advance in applying a unified model to different modalities.

\section{Public Dataset}
In this section, we introduce several image-report datasets used for AMRG. All discussed datasets are publicly available and privacy-safe, meaning they are free of any patient-identifying information. The two benchmark datasets (Section \ref{Benchmark Datasets}) are widely used and serve as standards for performance comparison in most AMRG studies. Additionally, other datasets (Section \ref{Other Datasets}) have been created to address specific needs, such as particular languages and imaging modalities.

\subsection{Benchmark Datasets}
\label{Benchmark Datasets}
\textbf{IU-Xray: }The Indiana University Chest X-ray dataset (IU-Xray) \cite{demner2016preparing}, also known as the OpenI dataset, was released in 2016. This dataset was sourced from two large hospital systems within the Indiana Network for Patient Care database. It comprises 7,470 chest radiographs (including both frontal and lateral views) and 3,955 corresponding narrative reports from 3,955 patients. Each report includes two primary sections: findings, which provide a detailed natural language description of the significant aspects in the image, and impression, which offer a concise summary of the most immediately relevant findings. The 3,955 studies are divided into 2,470 abnormal cases and 1,485 normal cases. Disease labels were extracted from the reports either manually or automatically using Medical Subject Headings (MeSH) \cite{fb1963medical}, Radiology Lexicon (RadLex) \cite{langlotz2006radlex}, and the Medical Text Indexer (MTI) \cite{mork2013nlm}. The ten most frequent disease tags are cardiomegaly, pulmonary atelectasis, calcified granuloma, tortuous aorta, hypoinflated lung, lung base opacity, pleural effusion, lung hyperinflation, lung cicatrix, and lung calcinosis. Since the dataset does not have an official split, the common practice is to randomly divide it into training, validation, and test sets in a 7:1:2 ratio.  

\textbf{MIMIC-CXR: }The Medical Information Mart for Intensive Care Chest X-ray (MIMIC-CXR) \cite{johnson2019mimic} is the largest public medical image-report dataset. It includes imaging studies from 65,379 patients from the Beth Israel Deaconess Medical Center Emergency Department, collected between 2011 and 2016. The dataset includes 377,110 chest radiographs (including frontal and lateral views) and 227,835 corresponding reports, most of which include findings and impression sections. Each report is associated with to one or more images. On average, 3.5 reports from different time periods are collected for each patient, providing longitudinal data that allows researchers to reference previous images. The dataset is officially split into training, validation, and test sets, which improves reproducibility. Specifically, the training set contains 368,960 images and 222,758 reports, the validation set contains 2,991 images and 1,808 reports, and the test set contains 5,159 images and 3,269 reports.

The images were originally stored in DICOM format, but a JPEG version (MIMIC-CXR-JPG) \cite{johnson2019mimicjpg} was also created to reduce storage size. In addition, 14 structured disease labels were extracted from the reports using NegBio \cite{peng2018negbio} and Chexpert \cite{irvin2019chexpert}, including atelectasis, cardiomegaly, consolidation, edema, enlarged cardiomediastinum, fracture, lung lesion, lung opacity, pleural effusion, pneumonia, pneumothorax, pleural other, support devices, and no finding.

Due to the large volume and the diversity of diseases in the MIMIC-CXR dataset, two derived datasets were created. MIMIC-ABM \cite{ni2020learning} is a subset that contains only abnormal studies with at least one abnormal finding. This subset, consisting of 38,551 pairs (26,946 for training, 3,801 for validation, and 7,804 for testing), addresses data bias caused by the prevalence of normal studies in the original dataset, allowing models to learn abnormal patterns more effectively. Another derived dataset, MIMIC-PRO \cite{ramesh2022improving}, eliminates all temporal information within the reports. Training with this dataset helps mitigate the generation of hallucinations about non-existent priors by the model.







\subsection{Other Datasets}
\label{Other Datasets}
In addition to the benchmark dataset of chest radiographs with English reports, there are several public datasets containing non-English reports and non-radiographic images. These datasets expand the range of languages and imaging modalities, catering to the specific needs of different research communities.

\textbf{Padchest: }The Pathology Detection in Chest Radiographs (Padchest) \cite{bustos2020padchest} contains imaging studies of 67,625 patients collected from 2009 to 2017 at Hospital San Juan, Spain. It contains 160,868 chest radiographs and 109,931 reports. The radiographs include six views: postero-anterior (PA), lateral, AP-horizontal, AP-vertical, costal, and pediatric. All reports are written in Spanish, and each report potentially corresponds to one or multiple views.

\textbf{CX-CHR: }CX-CHR \cite{li2018hybrid} contains 45,598 chest radiographs and 33,236 reports written in Chinese. Each report corresponds to one or multiple views, including PA and lateral views. The reports include findings and impression sections, as well as labels for 20 common chest diseases. Although this dataset is internally proprietary, researchers can apply for academic use after signing a confidentiality agreement, and it has been used in \cite{wang2020unifying} and \cite{li2023auxiliary}.

\textbf{COV-CTR: }The COVID-19 CT Report dataset (COV-CTR) \cite{li2023auxiliary} contains 728 lung CT scans and corresponding reports. The images are from the public COVID-CT dataset \cite{yang2020covid}, and the reports are written in Chinese by three radiologists from the First Affiliated Hospital of Harbin Medical University. Of these studies, 349 are COVID-19 cases, and 379 are non-COVID-19 cases.

\textbf{FFA-IR: }The Fundus Fluorescein Angiography Images and Reports (FFA-IR) \cite{li2021ffa} was collected from patients at the Zhongshan Ophthalmic Center of Sun Yat-Sen University in Guangzhou, China, between November 2016 and December 2019. The dataset comprises 1,048,584 FFA images and 10,790 reports, encompassing 46 categories of retinal lesions. Each report is bilingual, with both Chinese and English versions available. Approximately 5\% of the cases are healthy, while the remaining cases present various retinal conditions. The dataset is divided into official splits: 8,016 cases for training, 1,069 cases for validation, and 1,604 cases for testing, facilitating future model performance comparisons. Each case includes not only the report and FFA images but also explainable annotations to enhance the interpretability of AMRG models. Specifically, ophthalmologists labeled the lesion locations and categories in the images with rectangular boxes based on the size, location, and stage of the lesions described in the report.

\textbf{EndoVis-18: }The EndoVis-18 dataset originates from the MICCAI Robotic Scene Segmentation of Endoscopic Vision Challenge 2018 \cite{xu2021class,xu2021learning}. It comprises 1,560 endoscopic surgical images, each annotated by experienced surgeons \cite{xu2021class,xu2021learning}, and corresponding scene graphs are generated by \cite{islam2020learning}. The dataset includes a total of nine objects, featuring one type of tissue (kidney) and eight different surgical instruments. Additionally, there are 11 types of interactions between the surgical instruments and tissue, such as manipulation, grasping, and cutting. Following the methodology of previous studies, 1,124 images along with their captions and scene graphs are used as the training set, while the remaining images serve as the test set.

\textbf{TORS: }The TORS dataset was collected from transoral robotic surgery \cite{xu2021class,xu2021learning} and consists of 335 surgical images, also annotated by experienced surgeons with associated scene graphs. It includes five types of objects: tissue, clip applier, suction, spatulated monopolar cautery, and Maryland dissector, with eight types of semantic interactions such as clipping, suturing, and grasping. 
\section{Evaluation Metrics}
This section discusses various evaluation metrics used to evaluate the quality of generated reports, including (\rmnum{1}) NLP metrics (Section \ref{NLP metrics}), (\rmnum{2}) clinical efficacy (CE) metrics (Section \ref{Clinical Efficacy}), and (\rmnum{3}) human evaluation (Section \ref{Human Evaluation}). NLP metrics measure the word overlap between the generated report and the reference report, while CE metrics evaluate the clinical accuracy of the generated reports by focusing on specific disease labels. Human evaluation involves inviting radiologists to assess the generated reports to ensure reliability.

\subsection{NLP metrics}
\label{NLP metrics}
NLP metrics were originally designed for natural language tasks and are employed in AMRG tasks to measure the quality of generated reports. Commonly used NLP metrics include BLEU, METEOR, ROUGE-L, and CIDEr, which are described in detail in the following sections.

\textbf{BLEU: }Bilingual Evaluation Understudy (BLEU) \cite{papineni2002bleu} was originally designed for machine translation. It measures the correspondence between a candidate (generated) sequence and a reference (ground-truth) sequence. A higher BLEU score indicates a closer match to the reference sequence. 

BLEU calculates precision ($p_n$) for n-grams (subsequences of n words) and includes a brevity penalty ($BP$) for overly short sentences. BLEU is computed as follows:
\begin{equation}
\begin{aligned}
    p_n &= \frac{\text{Number of n-grams in candidate that match reference}}{\text{Total number of n-grams in candidate}}\\
    BP &= 
\begin{cases} 
1 & \text{if } c > r \\
e^{(1 - \frac{r}{c})} & \text{if } c \leq r
\end{cases}, \quad \text{BLEU} = BP \cdot \exp(\sum^N_{n=1}w_n\log p_n),
\end{aligned}
\end{equation}
where $c$ and $r$ are the lengths of the candidate and reference sequence, respectively, and $w_n$ is the weight for n-grams (typically $w_n = \frac{1}{N}$).

\textbf{METEOR: }The Metric for Evaluation of Translation with Explicit Ordering (METEOR) \cite{banerjee2005meteor} is a precision and recall-based measure that improves BLEU-1 by considering synonyms, stemming, and word order. It expands uni-gram matching to include exact matches, stemming matches, and synonym matches. METEOR aligns more closely with human judgment when comparing candidate and reference sequences. 

Precision ($P$) and recall ($R$) are computed as follows:
\begin{equation}
    P =\frac{m}{\text{length of candidate}}, \quad
    R =\frac{m}{\text{length of reference}}, 
\end{equation}
where $m$ is the number of matches (including exact, stemmed, and synonym matches) between candidate and reference uni-grams. 

The harmonic mean ($F_{mean}$) of precision and recall is:
\begin{equation}
    F_{mean} = \frac{10 P R}{R+ 9P}
\end{equation}

METEOR also includes a fragmentation penalty ($Pen$) to penalize candidate sequences with poor word order:
\begin{equation}
    Pen = 0.5 \cdot \frac{\# \text{chunks}}{m},
\end{equation}
where $\# \text{chunks}$ are the number of groups of matched words in the same order as the reference. The final METEOR score is formulated as:
\begin{equation}
    \text{METEOR} = F_{mean} \cdot (1-Pen)
\end{equation}

\textbf{ROUGE-L: } Recall-Oriented Understudy for Gisting Evaluation - Longest Common Subsequence (ROUGE-L) \cite{lin2004rouge} is a metric originally designed for automatic summarization. It measures the longest common subsequence (LCS) between the candidate ($X$) and reference ($Y$) sequences. A higher ROUGE-L score indicates better quality in terms of the structure and important content of the reference sequence. 

Precision ($P$) and recall ($R$) are computed by:
\begin{equation}
\centering
P = \frac{LCS(X,Y)}{\text{length of candidate}}, \quad R = \frac{LCS(X,Y)}{\text{length of reference}},
\end{equation}
where $LCS(X,Y)$ denotes the length of the LCS of sequences $X$ and $Y$. 

ROUGE-L is formulated as:
\begin{equation}
    \text{ROUGE-L} = \frac{(1+\beta^2)PR}{R+\beta^2 P},
\end{equation}
where $\beta$ is a hyper-parameter that determines the relative importance of precision and recall. It is usually set to a large number to emphasize the recall score.

\textbf{CIDEr: }Consensus-based Image Description Evaluation (CIDEr) \cite{vedantam2015cider} was designed for image captioning.  It measures the cosine similarity between a generated caption and a set of reference captions using the term frequency-inverse document frequency (TF-IDF) weighted n-grams. TF-IDF assigns higher weights to significant words, so a high CIDEr score indicates substantial coverage of these important words. TF quantifies how frequently an n-gram appears in a caption, while IDF measures how common an n-gram appears across all reference captions. The TF and IDF are  calculated as follows:
\begin{equation}
        \text{TF}_{ij} = \frac{\text{Count}_{ij}}{\sum_k \text{Count}_{ik}}, \quad
        \text{IDF}_i = \log \frac{N}{\sum_j \min (1, \text{Count}_{ij})},
\end{equation}
where $\text{Count}_{ij}$ is the count of the n-gram $i$ in caption $j$, $\sum_k \text{Count}_{ik}$ is the total count of all n-grams in caption $j$, $N$ is the total number of reference captions, and $\sum_j \min (1, \text{Count}_{ij})$ is the number of reference captions containing the n-gram $i$. 

The TF-IDF weighting for an n-gram $i$ in caption $j$ is calculated as:
\begin{equation}
    \text{TF-IDF}_{ij} = \text{TF}_{ij} \cdot  \text{IDF}_i
\end{equation} 

Denote TF-IDF vector for n-grams of length $n$ and caption $j$ as $v_{j,n}$. The $\text{CIDEr}_n$ score for n-grams of length $n$ is computed by averaging the cosine similarity between the candidate and reference captions over all reference captions:
\begin{equation}
    \text{CIDEr}_n (c,R) = \frac{1}{|R|}\sum_{r \in R} S (v_{c,n}, v_{r,n}),
\end{equation}
where $R$ is the set of reference captions, $|R|$ is the number of reference captions, $S$ represents cosine similarity, and $v_{c,n}$ and $v_{r,n}$ are TF-IDF vectors for the candidate and reference captions for n-grams of size $n$. 

The final CIDEr score is a weighted average of the $\text{CIDEr}_n$ scores for different n-gram lengths:
\begin{equation}
     \text{CIDEr} (c,R) = \frac{1}{N} \sum^N_{n=1} w_n \cdot \text{CIDEr}_n (c,R),
\end{equation}
where $N$ is the maximum n-gram length (typically 4), and $w_n$ is the weight for n-grams of length $n$ (typically $w_n = \frac{1}{N}$). 

\subsection{Clinical Efficacy}
\label{Clinical Efficacy}
NLP metrics primarily measure the word overlap between the generated report and the reference report. However, in the medical field, semantic similarity and factual consistency between the generated report and the reference report are more important. For example, the NLP scores for the sentences ``The heart is within normal size and contour'' and ``No cardiomegaly observed'' are zero. However, in medical reports, these two sentences convey the same information. Therefore, many radiographic report generation studies supplement NLP metrics with clinical efficacy (CE) metrics \cite{huang2023kiut, liu2021auto, liu2024bootstrapping, wang2023r2gengpt, yang2023radiology,    liu2023observation, zhang2023semi, endo2021retrieval, chen2020generating, yan2021weakly, miura2020improving, chen2022cross, hirsch2024medcycle, qin2022reinforced, hou2023organ, liu2024multi, wang2024camanet}. Specifically, they use CheXpert \cite{irvin2019chexpert} to extract labels from the generated and reference reports, focusing on 12 possible chest diseases. The label-based precision, recall, and F1 score are then calculated as CE metrics. This approach allows CE metrics to assess whether the generated report and the reference report contain the same diseases.

However, CE metrics are currently limited to the evaluation of English chest radiography reports. There are no tools similar to CheXpert for extracting disease labels from text for other body parts, modalities, and languages, which presents an opportunity for future research.

\subsection{Human Evaluation}
\label{Human Evaluation}
However, both NLP metrics and CE metrics sometimes are unreliable for evaluating medical reports, and CE metrics are limited to pre-trained disease categories. Some researchers suggest introducing human evaluation to comprehensively assess the quality of generated reports \cite{miura2020improving, qin2022reinforced, li2023auxiliary, li2021ffa, zhang2019optimizing, liu2022competence}. Specifically, reports generated by multiple candidate models are mixed together, and multiple board-certified radiologists compare the candidate reports with the reference reports to avoid personal bias. The radiologists select the generated reports that are most similar to the reference reports based on fluency, factual consistency, and overall quality. While human evaluation is the most reliable evaluation method, it is also the most expensive and impractical for large-scale evaluations.

\section{Performance Comparisons}
Table \ref{tab:results} shows the results of SOTA radiographic reporting methods published between 2021 and 2024 on benchmark radiography datasets. By comparing their performance and techniques, we identified six techniques that effectively improve NLP and CE metrics: (\rmnum{1}) human-computer interaction, (\rmnum{2}) reinforcement learning, (\rmnum{3}) detection and segmentation, (\rmnum{4}) disease classification, (\rmnum{5}) traceback mechanism, and (\rmnum{6}) local alignment. In the following sections, BLEU-4, METEOR, ROUGE-L, and CIDEr in NLP metrics, as well as precision, recall, and F1 score in CE metrics, are simplified to BL-4, MTR, RG-L, CD, P, R, and F, respectively.

\begin{figure}[htbp]
    \centering
    \includegraphics[width=1.0\textwidth]{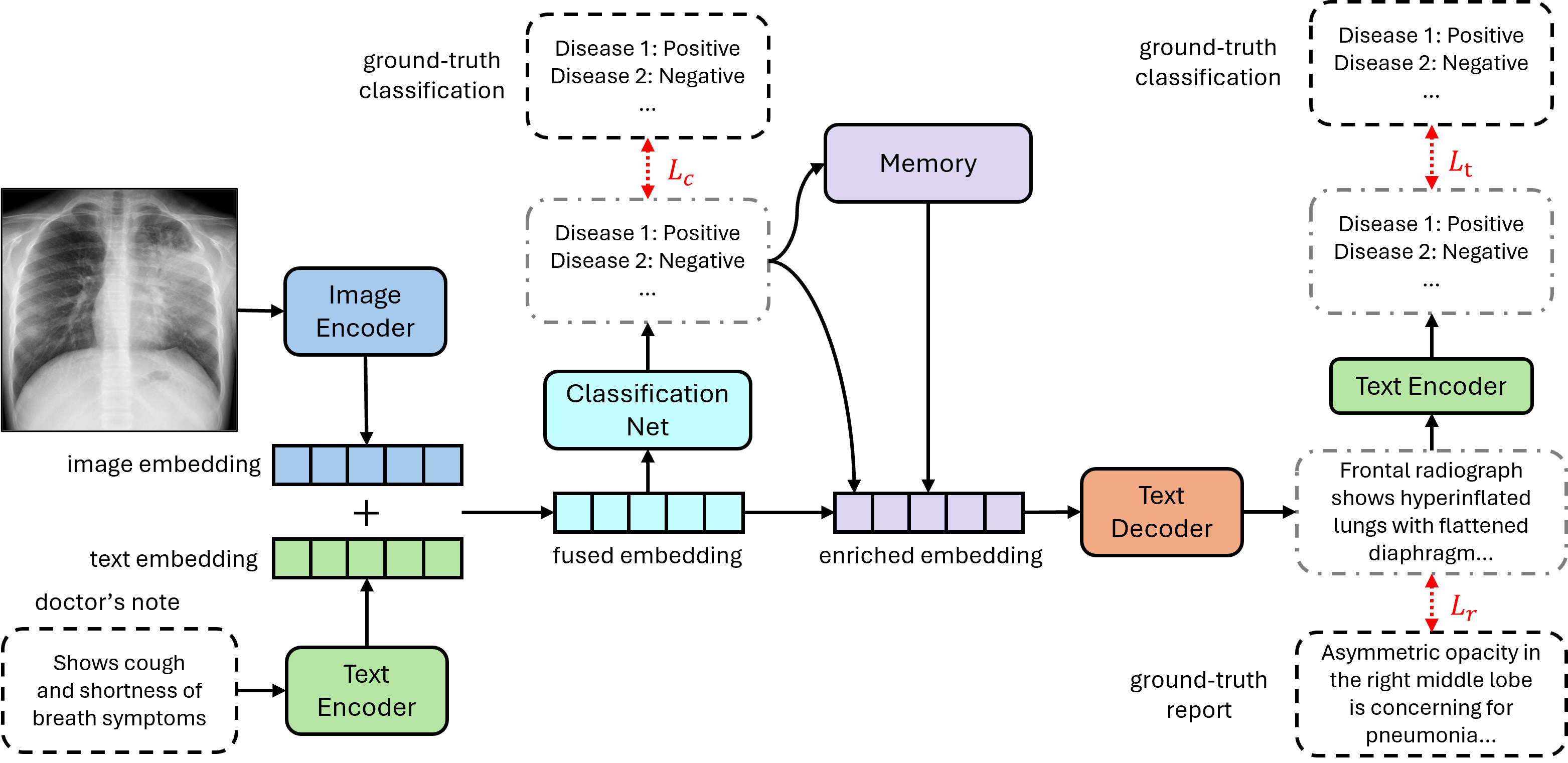}
    \caption{This diagram illustrates the architecture of the model developed by Nguyen et al. \cite{nguyen2021automated}, which involves human-computer interaction, disease classification, memory retrieval, and a traceback mechanism. The model begins by encoding a chest radiograph and the doctor’s note into image and text embeddings. These embeddings are combined to form a fused embedding, which is then processed by a classification network to predict the patient’s diseases. The predicted diseases guide a search through stored memory to retrieve relevant information. The fused embedding, predicted diseases, and retrieved memory are integrated to create an enriched embedding, which is subsequently decoded into a report. This generated report is further classified using a text encoder-based classifier to verify whether the diseases identified in the report align with the diseases indicated by the image. Model optimization is driven by three loss functions: the classification loss ($\mathcal{L}_c$) between the predicted and ground truth (GT) diseases, the report loss ($\mathcal{L}_r$) between the generated and GT reports, and the traceback loss ($\mathcal{L}_t$) between the diseases in the generated report and the GT diseases. The total loss is $\mathcal{L}_{total} = \mathcal{L}_c + \mathcal{L}_r + \mathcal{L}_t$.}
    \label{human}
\end{figure}

Human-computer interaction technology achieves the highest scores in both NLP and CE. Specifically, the inclusion of doctors' notes significantly enhances the quality of the generated reports. Nguyen et al. \cite{nguyen2021automated} achieved the best NLP scores on both MIMIC-CXR (BL-4: 0.224, MTR: 0.222, and RG-L: 0.390) and IU-Xray (BL-4: 0.235, MTR: 0.219, and RG-L: 0.436) by incorporating clinical documents. These clinical documents, which may include patients’ clinical histories or doctors’ notes, guide the model to focus on specific areas of the image and relevant diseases. For instance, if the doctor's note mentions ``shows cough and shortness of breath symptoms'', the model will focus on the lung area and consider pneumonia.  In addition to human-computer interaction, the model involves disease classification, memory retrieval and traceback mechanism, as illustrated in Figure \ref{human}. Similarly, Liu et al. \cite{liu2023observation} introduced disease labels provided by radiologists, achieving the highest CE scores (P: 0.855, R: 0.730, and F: 0.773). Their model offers two options: automatic disease classification based on the input image or radiologist-provided potential disease labels. The latter results in markedly higher clinical efficacy in generated reports. Thus, incorporating human guidance into the model effectively improves the quality of the generated reports.

For methods without human interaction, reinforcement learning (Section \ref{Reinforcement Learning}) is most effective. Xu et al. \cite{xu2023hybrid} used BL-4, MTR, and CD as rewards and achieved the second-highest NLP scores (BL-4: 0.192, MTR: 0.207, and RG-L: 0.380) on the MIMIC-CXR dataset. Likewise, the factual completeness and consistency reward designed by Miura et al. \cite{miura2020improving} resulted in the second-highest CE scores (P: 0.503, R: 0.651, and F: 0.567) on MIMIC-CXR dataset and highest CD scores (CD: 0.509 and 1.034) on both datasets. In addition, using CD alone as a reward can also lead to relatively high NLP and CE scores on benchmark datasets \cite{wang2021self, liu2024multi, wang2022medical}. Therefore, incorporating reinforcement learning into AMRG models is a straightforward but effective strategy.

\begin{figure}[htbp]
    \centering
    \includegraphics[width=1.0\textwidth]{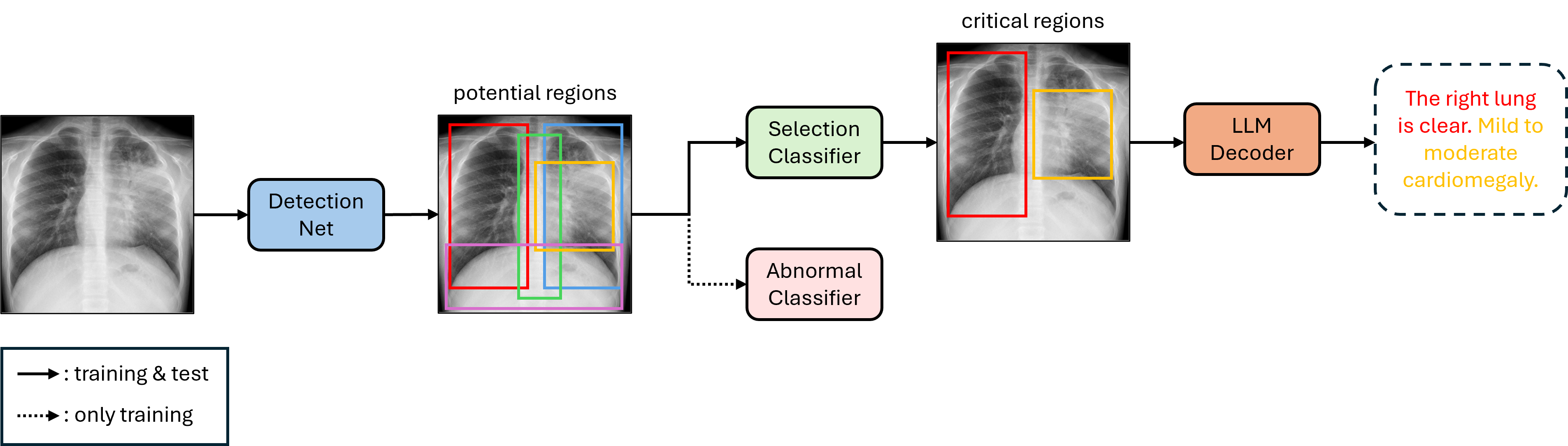}
    \caption{This diagram illustrates the architecture of the model developed by Tanida et al. \cite{tanida2023interactive}, which integrates detection, disease classification, and a large language model (LLM). The detection network extracts visual features from 29 potential anatomical regions in chest radiographs. These features are then processed by an abnormal classifier and a selection classifier. The abnormal classifier determines whether a region contains a lesion, encoding strong abnormal information into the features. The selection classifier identifies regions critical for report generation, ensuring that only the visual features from these critical regions are passed to the decoder. The model's decoder, which incorporates a pre-trained LLM, injects the features of the selected critical regions through pseudo self-attention, generating a sentence for each region.}
    \label{detection_figure}
\end{figure}

Another effective technique involves leveraging pre-trained detection or segmentation networks (Section \ref{Detection and Segmentation}) to enhance AMRG models by focusing on meaningful anatomical regions. For instance, Tanida et al. \cite{tanida2023interactive} integrated a detection network with binary classifiers, as illustrated in Figure \ref{detection_figure}, enabling the model to concentrate on critical regions and achieving the second-highest CD score of 0.495 on the MIMIC-CXR dataset. Similarly, Zhao et al. \cite{zhao2023medical} employed a segmentation network, securing the second-highest NLP scores (BL-4: 0.221 and RG-L: 0.433) on the IU-Xray dataset.

In addition, integrating multi-label disease classification (Section \ref{Disease Classification}) yields considerable improvements in CE scores \cite{jin2024promptmrg, hou2023organ, yang2023radiology, hou2023recap, liu2021auto, huang2023kiut}, ensuring that the diseases identified in the generated reports are consistent with those in the input images. For implementation, the image encoder can use disease classification as a pre-training task, or the classification can be employed as a joint learning task during training. Notably, incorporating the classification results as additional information into the text decoder can yield higher CE metrics. For example, Li et al. \cite{jin2024promptmrg} reported CE scores of P: 0.501, R: 0.509, and F: 0.476, while Hou et al. \cite{hou2023organ} achieved CE scores of P: 0.416, R: 0.418, and F: 0.385.

For the IU-Xray dataset, both the traceback mechanism (Section \ref{Traceback Mechanism}) \cite{li2022self, wang2021self, ye2024dual} and local alignment (Section \ref{Local Alignment}) \cite{wang2023fine, liu2024multi, ye2024dual} demonstrate notable efficacy. The traceback mechanism involves making the generated report similar to the reference report at the feature level, while local alignment aligns sentences or words with image patches. In particular, Li et al. \cite{li2022self} employed a traceback mechanism to achieve notable NLP scores (BL-4: 0.215,and RG-L: 0.415), indicating that the generated reports are similar to the reference reports in terms of 4-gram and the longest sequence. Ye et al. \cite{ye2024dual} combined traceback with local alignment and obtained the highest MTR score of 0.233 and a relative high CD score of 0.469, indicating that the generated reports have a high overlap with reference reports in terms of keywords. However, these techniques did not yield similarly excellent results on the MIMIC-CXR dataset, indicating that the traceback mechanism and local alignment still have limitations when applied to more complex and variable datasets.

\begin{table}[htbp!]
\caption{Comparison of radiographic report generation models (2021–2024) on benchmark datasets. The `Code' column shows if the code is publicly accessible, and the `Method' column lists the techniques used. Techniques mentioned in Section 3 follow the names in Figure 2, while others retain their original names (e.g., curriculum learning). The highest and second-highest values in each column are bolded and underlined, respectively. Abbreviations: BL-4 (BLEU-4), MTR (METEOR), RG-L (ROUGE-L), CD (CIDEr), P (Precision), R (Recall), and F (F1 Score).}
\label{tab:results}
\centering
\renewcommand{\arraystretch}{1.1}
\rowcolors{3}{gray!15}{white}
\resizebox{\textwidth}{!}{
\begin{tabular}{m{3.3 cm} 
>{\centering\arraybackslash}m{1cm} 
>{\centering\arraybackslash}m{1cm} |
>{\centering\arraybackslash}m{1.4cm} 
>{\centering\arraybackslash}m{1.4cm} 
>{\centering\arraybackslash}m{1.4cm} 
>{\centering\arraybackslash}m{1.4cm} 
>{\centering\arraybackslash}m{1.4cm} 
>{\centering\arraybackslash}m{1.4cm} 
>{\centering\arraybackslash}m{1.4cm} |
>{\centering\arraybackslash}m{1.4cm} 
>{\centering\arraybackslash}m{1.4cm} 
>{\centering\arraybackslash}m{1.4cm} 
>{\centering\arraybackslash}m{1.4cm} |
m{9cm} |
>{\centering\arraybackslash}m{2.5cm}
}
\multicolumn{1}{c}{\textbf{Model}} & \textbf{Year} & \textbf{Code} & \multicolumn{7}{c|}{\textbf{MIMIC-CXR}} & \multicolumn{4}{c|}{\textbf{IU-Xray}} & \multicolumn{1}{c}{\textbf{Method}} & \textbf{Dataset} \\ 
& & & \textbf{BL-4} & \textbf{MTR} & \textbf{RG-L} & \textbf{CD} & \textbf{P} & \textbf{R} & \textbf{F} & \textbf{BL-4} & \textbf{MTR} & \textbf{RG-L} & \textbf{CD} & & \\ \hline
Liu et al. \cite{liu2021exploring} & 2021 &  &
0.106 & 0.149 & 0.284 & --
& -- & -- & -- &
0.168 & -- & 0.376 & 0.351 & Encoder structure, retrieve similarity reports, knowledge graph& 
MIMIC-CXR, IU-Xray\\

Yan et al. \cite{yan2021weakly} & 2021 & &
0.107 & 0.144 & 0.274 & --
& 0.385 & 0.274 & 0.294 &
 -- & -- &  -- &  -- & Global alignment& 
MIMIC-CXR, MIMIC-ABM\\

Liu et al. \cite{liu2021contrastive} & 2021 & &
0.109 & 0.151 & 0.283 & --
& 0.352 & 0.298 & 0.303 &
 0.169 & 0.193 &  0.381 &  -- & Contrastive attention&
MIMIC-CXR, IU-Xray\\

You et al. \cite{you2021aligntransformer} & 2021 & &
0.112 & 0.158 & 0.283 & --
& -- & -- & -- &
 0.173 & 0.204 &  0.379 &  -- & Disease classification, encoder structure&
MIMIC-CXR, IU-Xray\\

Miura et al. \cite{miura2020improving} & 2021 & $\checkmark$ &
0.114 & -- & -- & \textbf{0.509}
& \underline{0.503} & \underline{0.651} & \underline{0.567} &
 0.131 & -- &  -- &  \textbf{1.034} & Reinforcement learning&
MIMIC-CXR, IU-Xray\\

Liu et al. \cite{liu2021auto} & 2021 &  &
0.118 &  0.153 & 0.295 & --
& 0.389 & 0.362 & 0.355 &
 0.179 & 0.195 &  0.383 & -- & 
Disease classification, memory, knowledge graph, unsupervised learning&
MIMIC-CXR, IU-Xray\\

Yang et al. \cite{yang2021joint} & 2021 & $\checkmark$ &
0.143 & -- & 0.326 & 0.273
& 0.237 & 0.326 & -- &
0.180 & -- & 0.398 & 0.439 & Global alignment, disease classification &
MIMIC-CXR, IU-Xray\\

Nguyen et al. \cite{nguyen2021automated} & 2021 & $\checkmark$ &
\textbf{0.224} & \textbf{0.222} & \textbf{0.390} & --
& 0.432 & 0.418 & 0.412 &
\textbf{0.235} & \underline{0.219} & \textbf{0.436} & -- & 
Human-computer interaction, disease classification, memory, traceback mechanism&
MIMIC-CXR, IU-Xray\\

Wang et al. \cite{wang2021self} & 2021 &  &
-- &  -- & -- & --
& -- & -- & -- &
 0.208 & -- &  0.359 & 0.452 & Global alignment, traceback mechanism, reinforcement learning&
COV-CTR, IU-Xray\\

Hou et al. \cite{hou2021ratchet} & 2021 & $\checkmark$ &
-- &  0.101 & 0.240 & 0.493
& -- & -- & -- &
 -- & -- &  -- & -- & Encoder structure&
MIMIC-CXR, IU-Xray\\

Alfarghaly et al. \cite{alfarghaly2021automated} & 2021 & $\checkmark$ &
-- &  -- & -- & --
& -- & -- & -- &
 0.111 & 0.164 &  0.289 & 0.257 & Large language model&
MIMIC-CXR, IU-Xray\\

Liu et al. \cite{liu2022competence} & 2022 & &
0.097 &  0.133 & 0.281 & --
& -- & -- & -- &
 0.162 & 0.186 &  0.378 & -- &  Curriculum learning &
MIMIC-CXR, IU-Xray\\

Wang et al. \cite{wang2022cross} & 2022 & $\checkmark$ &
0.105 &  0.138 & 0.279 & --
& -- & -- & -- &
 0.199 & 0.22 &  0.411 & 0.359 & Intermediate matrix &
MIMIC-CXR, IU-Xray\\

Chen et al. \cite{chen2022cross} & 2022 & $\checkmark$ &
0.106 &  0.142 & 0.278 & --
& 0.334 & 0.275 & 0.278 &
 0.170 & 0.191 &  0.375 & -- & Intermediate matrix&
MIMIC-CXR, IU-Xray\\

Qin et al. \cite{qin2022reinforced} & 2022 & $\checkmark$ &
0.109 &  0.151 & 0.287 & --
& 0.342 & 0.294 & 0.292 &
0.181 & 0.201 &  0.384 & -- & Intermediate matrix, reinforcement learning &
MIMIC-CXR, IU-Xray\\

Yang et al. \cite{yang2022knowledge} & 2022 & $\checkmark$ &
0.115 & -- & 0.284 & 0.203
& -- & -- & -- &
0.178 & -- &  0.381 & 0.382 & Knowledge graph &
MIMIC-CXR, IU-Xray\\

Wang et al. \cite{wang2022automated} & 2022 & &
0.118 & -- & 0.287 & 0.281
& -- & -- & -- &
0.175 & -- &  0.377 & 0.449 & Global alignment, disease classification &
MIMIC-CXR, IU-Xray\\

Wang et al. \cite{wang2022inclusive} & 2022 & $\checkmark$ &
0.121 & 0.147 & 0.284 & --
& -- & -- & -- &
0.188 & 0.208 &  0.382 & -- & Disease classification &
MIMIC-CXR, IU-Xray\\

Wang et al. \cite{wang2022medical} & 2022 & $\checkmark$ &
0.136 & 0.170 & 0.298 & 0.429
& -- & -- & -- &
-- & -- &  -- & -- & Disease classification, encoder structure, memory, reinforcement learning &
MIMIC-CXR\\

Najdenkoska et al. \cite{najdenkoska2022uncertainty} & 2022 & $\checkmark$ &
0.136 & 0.191 & 0.315 & --
& 0.396 & 0.312 & 0.350 &
0.170 & 0.230 &  0.390 & -- & Global alignment &
MIMIC-CXR, IU-Xray\\

You et al. \cite{you2022jpg} & 2022 & &
-- & -- & -- & --
& -- & -- & -- &
0.174 & 0.193 &  0.377 & -- & Intermediate matrix, disease classification&
IU-Xray\\

Wang et al. \cite{wang2020unifying} & 2022 & &
-- & -- & -- & --
& -- & -- & -- &
0.175 & -- &  0.36 & 0.331 & Disease classification &
CX-CHR, IU-Xray\\

Li et al. \cite{li2022self} & 2022 & &
-- & -- & -- & --
& -- & -- & -- &
0.215 & 0.201 &  0.415 & -- & Traceback mechanism &
IU-Xray\\

Li et al. \cite{li2023auxiliary} & 2023 & $\checkmark$  &
-- & -- & -- & --
& -- & -- & -- &
0.125 & -- &  0.279 & 0.306 & Disease classification, detection, knowledge graph &
 CX-CHR, COV-CTR, IU-Xray\\

Lu et al. \cite{lu2023effectively} & 2023 &  &
0.069 & -- & 0.235 & --
& -- & -- & 0.32 &
-- & -- &  -- & -- & Large language model &
 MIMIC-CXR\\

Bannur et al. \cite{bannur2023learning} & 2023 & $\checkmark$ &
0.092 & -- & 0.296 & --
& -- & -- & -- &
-- & -- &  -- & -- & Global alignment, local alignment, eliminate noise &
 MIMIC-CXR, MS-CXR-T\\

Wu et al. \cite{wu2023multimodal} & 2023 & &
0.103 & 0.139 & 0.270 & 0.109
& -- & -- & -- &
0.18 & 0.206 &  0.369 & 0.287 & Global alignment &
MIMIC-CXR, IU-Xray\\

Li et al. \cite{li2023unify} & 2023 & &
0.107 & 0.157 & 0.289 & 0.246
& -- & -- & -- &
0.200 & 0.218 &  0.405 & 0.501 & Global alignment, intermediate matrix &
MIMIC-CXR, IU-Xray\\

Li et al. \cite{li2023dynamic} & 2023 & $\checkmark$ &
0.109 & 0.150 & 0.284 & 0.281
& -- & -- & -- &
0.163 & 0.193 &  0.383 & 0.586 & Global alignment, knowledge graph, eliminate noise &
MIMIC-CXR, IU-Xray\\

Yang et al. \cite{yang2023radiology} & 2023 & $\checkmark$ &
0.111 & -- & 0.274 & 0.111
& 0.420 & 0.339 & 0.352 &
0.174 &  -- &  0.399 & 0.407 & Disease classification, memory &
MIMIC-CXR, IU-Xray\\

Zhang et al. \cite{zhang2023novel} & 2023 & &
0.113 & 0.143 & 0.276 & --
& -- & -- & -- &
0.190 &  0.207 &  0.394 & -- & Memory &
MIMIC-CXR, IU-Xray, COV-CTR\\

Huang et al. \cite{huang2023kiut} & 2023 & &
0.113 & 0.160 & 0.285 & --
& 0.371 & 0.318 & 0.321 &
0.185 &  0.242 &  0.409 & -- & Disease classification, knowledge graph &
MIMIC-CXR, IU-Xray\\

Cao et al. \cite{cao2023mmtn} & 2023 & &
0.116 & 0.161 & 0.283 & --
& -- & -- & -- &
0.175 & -- &  0.375 & 0.361 & Encoder structure, memory &
MIMIC-CXR, IU-Xray\\

Wang et al. \cite{wang2023self} & 2023 & &
0.118 & 0.136 & 0.301 & --
& -- & -- & -- &
0.176 & 0.205 & 0.396 & -- & Detection &
MIMIC-CXR, IU-Xray\\

Wang et al. \cite{wang2023fine} & 2023 & &
0.119 & 0.158 & 0.286 & 0.259
& -- & -- & -- &
0.205 & 0.223 & 0.414 & 0.370 & Local alignment &
MIMIC-CXR, IU-Xray\\

Hou et al. \cite{hou2023organ} & 2023 & $\checkmark$ &
0.123 & 0.162 & 0.293 & --
& 0.416 & 0.418 & 0.385 &
0.195 & 0.205 & 0.399 & -- & Disease classification & 
MIMIC-CXR, IU-Xray\\

Wang et al. \cite{wang2023metransformer} & 2023 & &
0.124 & 0.152 & 0.291 & 0.362
& -- & -- & -- &
0.172 & 0.192 & 0.380 & 0.435 & Encoder structure &
MIMIC-CXR, IU-Xray\\

Hou et al. \cite{hou2023recap} & 2023 & $\checkmark$ &
0.125 & 0.168 & 0.288 & --
& 0.389 & 0.443 & 0.393 &
-- & -- & -- & -- & Disease classification, eliminate noise &
MIMIC-CXR, MIMIC-ABN\\

Tanida et al. \cite{tanida2023interactive} & 2023 & $\checkmark$ &
0.126 & 0.168 & 0.264 & \underline{0.495}
& -- & -- & -- &
-- & -- & -- & -- & Detection, disease classification, large language model &
MIMIC-CXR\\

Nicolson et al. \cite{nicolson2023improving} & 2023 & $\checkmark$ &
0.127 & 0.155 & 0.286 & 0.389
& 0.367 & 0.418 & 0.391 &
0.175 & 0.200 & 0.376 & \underline{0.694} & Warm starting &
MIMIC-CXR, IU-Xray\\

Si et al. \cite{si2023non} & 2023 & &
0.130 & 0.148 & 0.315 & --
& -- & -- & -- &
0.174 & -- & 0.388 & -- & Encoder structure, memory &
MIMIC-CXR, IU-Xray\\

Xu et al. \cite{xu2023hybrid} & 2023 & &
\underline{0.192} & \underline{0.207} & \underline{0.380} & 0.372
& -- & -- & -- &
0.149 & 0.197 & 0.381 &0.524 & Encoder structure, reinforcement learning &
MIMIC-CXR, IU-Xray\\

Wang et al. \cite{wang2023r2gengpt} & 2023 & $\checkmark$ &
0.134 & 0.160 & 0.297 & 0.269
& 0.392 & 0.387 & 0.389 &
0.173 & 0.211 & 0.377 & 0.438 & Large language model &
MIMIC-CXR, IU-Xray\\

Zhang et al. \cite{zhang2023semi} & 2023 & &
0.113 & 0.153 & 0.284 & --
& 0.380 & 0.342 & 0.335 &
0.165 & 0.195 & 0.377 & -- & Knowledge graph, semi-supervised learning &
MIMIC-CXR, IU-Xray\\

Liu et al. \cite{liu2023observation} & 2023 & &
0.125 & 0.160 & 0.304 & --
& \textbf{0.855} & \textbf{0.730} & \textbf{0.773} &
0.206 & 0.211 & 0.423 & -- & Human-computer interaction, global alignment, disease classification &
MIMIC-CXR, IU-Xray\\

Yan et al. \cite{yan2023attributed} & 2023 & &
-- & -- & 0.225 & 0.160
& -- & -- & -- &
-- & -- & 0.341 & 0.380 & Knowledge graph, disease classification &
MIMIC-CXR, IU-Xray\\

Zhao et al. \cite{zhao2023medical} & 2023 & &
-- & -- & -- & --
& -- & -- & -- &
\underline{0.221} & 0.210 & \underline{0.433} & -- & Global alignment, segmentation &
IU-Xray\\

Wang et al. \cite{wang2023mvco} & 2023 & &
-- & -- & -- & --
& -- & -- & -- &
0.157 & 0.196 & 0.374 & -- & Encoder's structure &
IU-Xray\\

Chen et al. \cite{chen2023fine} & 2024 & &
0.106 & 0.163 & 0.286 & --
& -- & -- & -- &
0.145 & 0.162 & 0.366 & -- & Encoder's structure, reinforcement learning &
MIMIC-CXR, IU-Xray\\

Jin et al. \cite{jin2024promptmrg} & 2024 & $\checkmark$ &
0.112 & 0.157 & 0.268 & --
& 0.501 & 0.509 & 0.476 &
0.098 & 0.160 & 0.281 & -- & Retrieve similarity reports, disease classification &
MIMIC-CXR, IU-Xray\\

Wang et al. \cite{wang2024camanet} & 2024 & $\checkmark$ &
0.112 & 0.145 & 0.279 & 0.161
& 0.483 &0.323 & 0.387 &
0.218 & 0.203 & 0.404 & 0.418 & Disease classification &
MIMIC-CXR, IU-Xray\\

Tu et al. \cite{tu2024towards} & 2024 & &
0.115 & -- & 0.275 & --
& -- & -- & 0.398 &
-- &  -- &  -- & -- & Disease classification &
MIMIC-CXR\\

Gao et al. \cite{gao2024simulating} & 2024 & $\checkmark$ &
0.116 & 0.168 & 0.286 & --
& 0.482 & 0.563 & 0.519 &
0.205 & 0.210 & 0.409 & -- & Retrieve similarity reports, memony &
MIMIC-CXR, IU-Xray\\

Yi et al. \cite{yi2024tsget} & 2024 & $\checkmark$ &
0.121 & 0.149 & 0.281 & --
& 0.319 & 0.509 & 0.393 &
0.194 & 0.218 & 0.402 & -- & Reinforcement learning &
MIMIC-CXR, IU-Xray\\

Pan et al. \cite{pan2023s3} & 2024 & &
0.125 & 0.154 & 0.291 & --
& -- & -- & -- &
0.172 & 0.206 & 0.401 & -- & Intermediate matrix &
MIMIC-CXR, IU-Xray\\

Liu et al. \cite{liu2024bootstrapping} & 2024 & $\checkmark$ &
0.128 & 0.175 & 0.291 & --
& 0.465 & 0.482 & 0.473 &
0.184 & 0.208 & 0.390 & -- & Retrieve similarity reports, large language model &
MIMIC-CXR, IU-Xray\\

Ye et al. \cite{ye2024dual} & 2024 & &
0.129 & 0.162 & 0.309 & 0.311
& -- & -- & -- &
0.204 & \textbf{0.233} & 0.386 & 0.469 & Local alignment, disease classification, traceback mechanism &
MIMIC-CXR, IU-Xray\\

Liu et al. \cite{liu2024multi} & 2024 & &
0.141 & 0.163 & 0.309 & --
& 0.457 & 0.337 & 0.330 &
0.175 & 0.192 & 0.379 & 0.368 & Local alignment, reinforcement learning &
MIMIC-CXR, IU-Xray\\

Hirsch et al. \cite{hirsch2024medcycle} & 2024 & &
0.072 & 0.128 & 0.239 & --
& 0.237 & 0.197 & 0.183 &
0.140 & 0.197 & 0.360 & -- & Unsupervised learning &
MIMIC-CXR, IU-Xray, PadChest\\

\hline
\end{tabular}}
\end{table}

\section{Future Directions}
Finally, we highlight unresolved issues in the current methods that present opportunities for future research in the AMRG field.

\textbf{Multimodal Learning: }The primary challenge in the AMRG field is bridging the modality gap between images and text. The CLIP model \cite{radford2021learning}, which utilizes natural language as supervision for contrastive learning, has significantly advanced this area. However, from the current results (Table \ref{tab:results}), the performance of SOTA AMRG models remains limited, with CIDEr scores much lower than those of SOTA general image captioning models \cite{li2023blip, li2022blip}. This discrepancy underscores the inadequacy of existing multimodal learning methods in fully supporting report generation models, particularly when dealing with medical images containing subtle differences.

One promising direction involves implementing local alignment techniques that associate specific image regions with textual entities, enabling the model to learn more fine-grained details. Although current local alignment methods have shown improvements on the IU-Xray dataset, they have not yielded significant benefits on more complex datasets such as MIMIC-CXR \cite{wang2023fine, liu2024multi, ye2024dual}. This indicates that current fine-grained alignment methods are still insufficient for the medical domain. Therefore, future research should focus on advancing these methods to capture the nuanced and subtle features of medical images more effectively.

\textbf{Unsupervised/Semi-Supervised Learning: }Another major factor limiting the AMRG models is the relatively small size of paired medical datasets. For instance, the largest medical dataset, MIMIC-CXR \cite{johnson2019mimic}, contains only 0.22 million pairs, whereas general image captioning datasets like Conceptual \cite{changpinyo2021conceptual} contain 12 million pairs. The high cost of creating image-text paired medical datasets makes expanding them to a similar scale as general datasets impractical. 

One potential solution is to employ unsupervised and semi-supervised learning to expand the available data. Some researchers have used image classification and text reconstruction to train image encoder and text decoder separately, allowing the model to learn valuable patterns and representations from unpaired data \cite{hirsch2024medcycle, liu2021auto, wang2022medclip}. This approach enables the use of image-only and text-only medical data to augment the training set. However, two main limitations persist: the need for disease labels for images and the low precision of current methods. Image classification for training encoder requires images with disease labels, which is also labor-intensive. Moreover, as shown in Table \ref{tab:results}, the performance of unsupervised \cite{hirsch2024medcycle, liu2021auto} and semi-supervised \cite{zhang2023semi} methods is currently lower than that of supervised methods. Future research should focus on eliminating the need for image labels and improving the performance of unsupervised and semi-supervised methods. By using larger datasets than those used in supervised methods, their accuracy could ultimately surpass that of supervised methods.

\textbf{Human-Computer Interaction: }Given the limitations of current methods in terms of accuracy, human-computer interaction systems represent a viable avenue for further development. In such systems, physicians can provide prompts to guide the model in generating descriptive reports \cite{nguyen2021automated, liu2023observation}, or the model can generate draft reports that physicians subsequently modify. Integrating the report generation model into the clinical diagnosis process can reduce repetitive tasks for physicians, allowing them to focus on diagnosing complex diseases. With physicians' supervision, the issue of low accuracy in the generated reports becomes less of an obstacle to clinical application, as physicians can refine the generated text.

Another approach to human-computer interaction involves incorporating physicians' feedback into the model's iteration process. This approach offers a more precise and targeted supervision signal than the loss function. By continuously optimizing the report generation model based on physicians' feedback during daily use, the model can become more adept at addressing specific diseases.

\textbf{Interpretability: }Another challenge to the clinical application of the AMRG models is the opacity of their decision-making process. Providing visual and textual explanations can assist physicians in understanding the rationale behind specific diagnostic recommendations. A common approach is to utilize back-propagated gradients to highlight pertinent regions within the image \cite{selvaraju2017grad}. However, given that the output of the AMRG models is lengthy text, this approach is unsuitable. To enhance interpretability, Chen et al. \cite{chen2023fine} attempted to show rectangular boxes on the image to indicate areas where the model believed lesions occurred and corresponded to specific generated sentences. Unfortunately, their model produced many overlapping boxes, which did not clearly explain the decision-making process.

A future direction for improving interpretability involves refining visual explanation techniques to highlight critical regions more precisely. Additionally, since descriptions of diseases in the report are more critical than those of normal conditions, combining visual explanations with textual ones would be beneficial. This approach can highlight key words or sentences in the generated text and the corresponding regions in the input image. Such a combination of visual and textual interpretation is more suitable for the AMRG domain.

\textbf{Evaluation Metrics: }Developing more accurate evaluation metrics to assess the accuracy of generated reports is also a critical need in current research. Current NLP evaluation metrics primarily measure word similarity between generated and reference texts, but they fail to capture the clinical accuracy of reports. Meanwhile, CE metrics are limited to chest radiographic reports and fixed disease categories. While some studies introduce human evaluation, its high cost and lack of standardized criteria hinder large-scale implementation.

The AMRG field needs specialized evaluation metrics or methods that can assess the correctness of medical terminology and the accuracy of diagnoses. These metrics should be also applicable to various image modalities and diseases, as well as handle the inherent variability in medical diagnoses. Additionally, such metrics should be scalable, cost-effective, and standardized to enable consistent comparisons of model performance, similar to existing NLP metrics.

\section{Conclusion}
In conclusion, the field of AMRG has made significant strides in recent years, addressing critical challenges and enhancing the efficiency and accuracy of medical diagnoses. Our comprehensive review of AMRG methods from 2021 to 2024 highlights fourteen solutions for the four primary challenges: modality gap, visual deviations, text complexity, and dataset limitations. We also present AMRG applications across various imaging modalities, including radiography, CT scans, MRI, ultrasound, FFA, endoscopic imaging, surgical scenes, and WSI. In addition, our review underscores the importance of publicly available datasets and robust evaluation metrics in advancing AMRG research. Based on their performance on benchmark datasets, we identify six solutions that can significantly improve evaluation metrics.

Despite these advancements, the field continues to face ongoing challenges. Future research should focus on developing more effective multimodal learning algorithms, enhancing human-computer interaction, expanding available datasets, improving model interpretability, and refining evaluation metrics to ensure greater accuracy.

\printbibliography
\end{document}